\documentclass[11pt]{article}

\usepackage{acl}
\usepackage{times}
\usepackage{latexsym}
\usepackage[T1]{fontenc}
\usepackage[utf8]{inputenc}
\usepackage{microtype}
\usepackage{inconsolata}
\usepackage{graphicx}
\graphicspath{{figs/}}
\usepackage{paralist}
\usepackage{amsmath,amssymb}
\newcommand{\model}{MARS-Gov}
\usepackage[ruled,vlined,linesnumbered]{algorithm2e}
\usepackage{mathtools}
\usepackage{fvextra} 
\usepackage{tcolorbox}
\tcbuselibrary{skins, breakable}
\definecolor{hg1}{HTML}{F7FCF5} 
\definecolor{hg2}{HTML}{E5F5E0}
\definecolor{hg3}{HTML}{C7E9C0}
\definecolor{hg4}{HTML}{A1D99B}
\definecolor{hg5}{HTML}{74C476} 
\definecolor{hr1}{HTML}{FFF5F0} 
\definecolor{hr2}{HTML}{FFE0D2}
\definecolor{hr3}{HTML}{FFAB91} 
\definecolor{hr4}{HTML}{FF8A80} 
\definecolor{cat_ref}{gray}{0.92}      
\definecolor{cat_zero}{HTML}{E3F2FD}   
\definecolor{cat_agent}{HTML}{F3E5F5}  
\definecolor{cat_ours}{HTML}{FFF3E0}   
\definecolor{salmon}{HTML}{FF8A80}
\definecolor{cat_debate}{HTML}{FFF2CC} 
\newcommand{\ca}{\cellcolor{hg1}}
\newcommand{\cb}{\cellcolor{hg2}}
\newcommand{\cc}{\cellcolor{hg3}}
\newcommand{\cd}{\cellcolor{hg4}}
\newcommand{\ce}{\cellcolor{hg5}}
\newcommand{\ra}{\cellcolor{hr1}}
\newcommand{\rb}{\cellcolor{hr2}}
\newcommand{\rc}{\cellcolor{hr3}}
\newcommand{\rd}{\cellcolor{hr4}}

 \usepackage{booktabs}
 \usepackage{multirow}
 \usepackage{tabularx}
 \usepackage{threeparttable}
 \usepackage{colortbl}  
 \usepackage{xcolor}    
 \usepackage{arydshln}  
 \usepackage{float}
 \usepackage{placeins}
\definecolor{aliceblue}{rgb}{0.94, 0.97, 1.0}  
\definecolor{mygreen}{rgb}{0.92, 0.98, 0.92}   
\definecolor{myred}{rgb}{0.98, 0.92, 0.92}     
\definecolor{darkblue}{RGB}{30, 50, 120}       
\definecolor{blue}{RGB}{30, 50, 120} 
\definecolor{alertcol}{HTML}{A31D1D}
\definecolor{safecol}{HTML}{1B6E2E}
\definecolor{stdcol}{HTML}{00401A}
\definecolor{caseframe}{HTML}{5D2E2E} 
\definecolor{caseback}{HTML}{FAF9F6}   
\definecolor{casedashed}{HTML}{5D2E2E} 
\definecolor{alertcol}{HTML}{A31D1D}
\definecolor{promptframe}{HTML}{6B4E8A}
\definecolor{promptback}{HTML}{FBF8FF}
\definecolor{prompttitle}{HTML}{54306F}
\definecolor{promptaccent}{HTML}{7C3AED}
\definecolor{promptwarn}{HTML}{B45309}
\definecolor{dutchpromptframe}{HTML}{9A5B16}
\definecolor{dutchpromptback}{HTML}{FFF9F0}
\definecolor{dutchprompttitle}{HTML}{7A3E12}
\definecolor{dutchpromptaccent}{HTML}{B45309}
\definecolor{dutchpromptwarn}{HTML}{9A3412}
\newcommand{\std}[1]{\textcolor{stdcol}{\scriptsize{$\pm$#1}}}
\newcommand{\prompttag}[1]{\textcolor{promptaccent}{\textbf{#1}}}
\newcommand{\promptwarn}[1]{\textcolor{promptwarn}{\textbf{#1}}}

\newcommand{\pfield}[1]{\texttt{\{#1\}}}
\DeclareUnicodeCharacter{2163}{IV}
\usepackage{pifont}
\usepackage{url}
\usepackage{multicol}
\usepackage{etoc}
\usepackage{hyperref}  
\usepackage{multicol}   
\usepackage{caption}   
\usepackage{enumitem}

\newtcolorbox{tracecase}[1]{
    enhanced,
    breakable,
    colback=gray!8,
    colframe=darkblue,
    colbacktitle=darkblue,
    coltitle=white,
    fonttitle=\bfseries\large,
    fontupper=\small,
    title={#1},
    sharp corners,
    boxrule=0.8pt,
    titlerule=0pt,
    left=1.2mm,
    right=1.2mm,
    top=1.0mm,
    bottom=1.0mm,
    before skip=0.75em,
    after skip=0.75em,
    segmentation style={dashed, darkblue!35},
    drop shadow=black!18
}
\newtcolorbox{promptbox}[1]{
    enhanced,
    breakable,
    colback=promptback,
    colframe=promptframe,
    colbacktitle=prompttitle,
    coltitle=white,
    fonttitle=\bfseries,
    fontupper=\small,
    title={#1},
    sharp corners,
    boxrule=0.7pt,
    titlerule=0pt,
    left=1.2mm,
    right=1.2mm,
    top=0.5mm,
    bottom=0.5mm,
    before skip=0.35em,
    after skip=0.35em,
    segmentation style={dashed, promptframe!35},
    drop shadow=black!12
}
\newtcolorbox{dutchpromptbox}[1]{
    enhanced,
    breakable,
    colback=dutchpromptback,
    colframe=dutchpromptframe,
    colbacktitle=dutchprompttitle,
    coltitle=white,
    fonttitle=\bfseries,
    fontupper=\small,
    title={#1},
    sharp corners,
    boxrule=0.7pt,
    titlerule=0pt,
    left=1.2mm,
    right=1.2mm,
    top=0.5mm,
    bottom=0.5mm,
    before skip=0.35em,
    after skip=0.35em,
    segmentation style={dashed, dutchpromptframe!35},
    drop shadow=black!12,
    before upper={\let\prompttag\dutchprompttag\let\promptwarn\dutchpromptwarn},
    before lower={\let\prompttag\dutchprompttag\let\promptwarn\dutchpromptwarn}
}

\title{The ``10th Juror'': Open-Set Standpoint Screening \\ for Bureaucratic Bias Detection}

\author{
  \textbf{Yuchen Miao\textsuperscript{1}},
  \textbf{Zijun Wang\textsuperscript{1}},
  \textbf{Chang Han\textsuperscript{1}},\\
  \textbf{Yurui Shi\textsuperscript{3}},
  \textbf{Mingtai Zhang\textsuperscript{4}},
  \textbf{Siyang Xu\textsuperscript{2,*}}\\[0.4em]
  \textsuperscript{1}Sydney Smart Technology College, Northeastern University, China\\
  \textsuperscript{2}School of Computer and Communication Engineering,\\
  Northeastern University at Qinhuangdao, China\\
  \textsuperscript{3}Taiyuan University of Technology\\
  \textsuperscript{4}School of Resources, Environment and Materials, Guangxi University
}

\begin{document}
\maketitle
\begingroup
\renewcommand{\thefootnote}{*}
\footnotetext{Corresponding author.}
\endgroup

\begin{abstract}
Presupposing the boundaries of bias is itself a form of bias. We study closed-loop bias governance for Dutch government documents, where a system must detect biased language, ground decisions in legal and contextual evidence, rewrite problematic sentences when intervention is warranted, and verify that the rewrite mitigates harm without distorting meaning. Existing methods face three challenges: (i) discriminative classifiers capture surface regularities but lack normative grounding; (ii) zero-shot LLMs often adopt generic viewpoints and over-flag ambiguous administrative language; and (iii) fixed taxonomies inherit the Closed-World Assumption, missing emerging local targets. We propose \model{}, a standpoint-aware multi-agent framework that combines legal retrieval, open-set target screening, specialized jurors, conservative routing, and rewrite verification. When screening finds an uncovered group, \model{} instantiates a dynamic ``10th juror'' to deliberate outside the fixed panel. On DGDB, \model{} sets a new SOTA with 0.880 F1, outperforming the strongest zero-shot LLM detector by 20.2 points (29.8\% relative) and the best supervised Dutch encoder by 6.8 points, while reducing unnecessary interventions to 2.5\%. Leave-One-Category-Out (LOCO) evaluation recovers held-out categories with 85.1\% Correct@1 and 93.8\% Correct@3.
\end{abstract}

\section{Introduction}

Government documents are a primary interface between the state and its citizens. Their language therefore does more than transmit information: it marks who is treated as credible, risky, dependent, or entitled to care. Linguistic neutrality matters for institutional trust \citep{rizk-lindgren-2025-adm, eubanks-2018-automating}, yet bias can persist even in restrained administrative prose \citep{deswart-etal-2025-dgdb, benjamin-2019-race}. It often appears not as overt toxicity \citep{waseem-hovy-2016-hateful}, but as exclusionary naming, negative framing, presupposition, or uneven standards of responsibility \citep{otmakhova-etal-2024-media}. A sentence may be formally polite and still attach suspicion, incompetence, or burden to a protected or socio-economically vulnerable group.

This setting makes bias detection different from ordinary harmful-language classification. The relevant question is not only whether a phrase sounds negative, but whether it is normatively grounded, legally salient, and administratively faithful. A reliable system must identify the affected standpoint, distinguish a necessary eligibility condition from a stigmatizing proxy, and avoid rewriting official text in ways that change factual obligations. These requirements turn detection into a governance problem: the system must explain, intervene only when warranted, and verify that any intervention preserves meaning.

\begin{figure}[t]
    \centering
    \includegraphics[width=\linewidth]{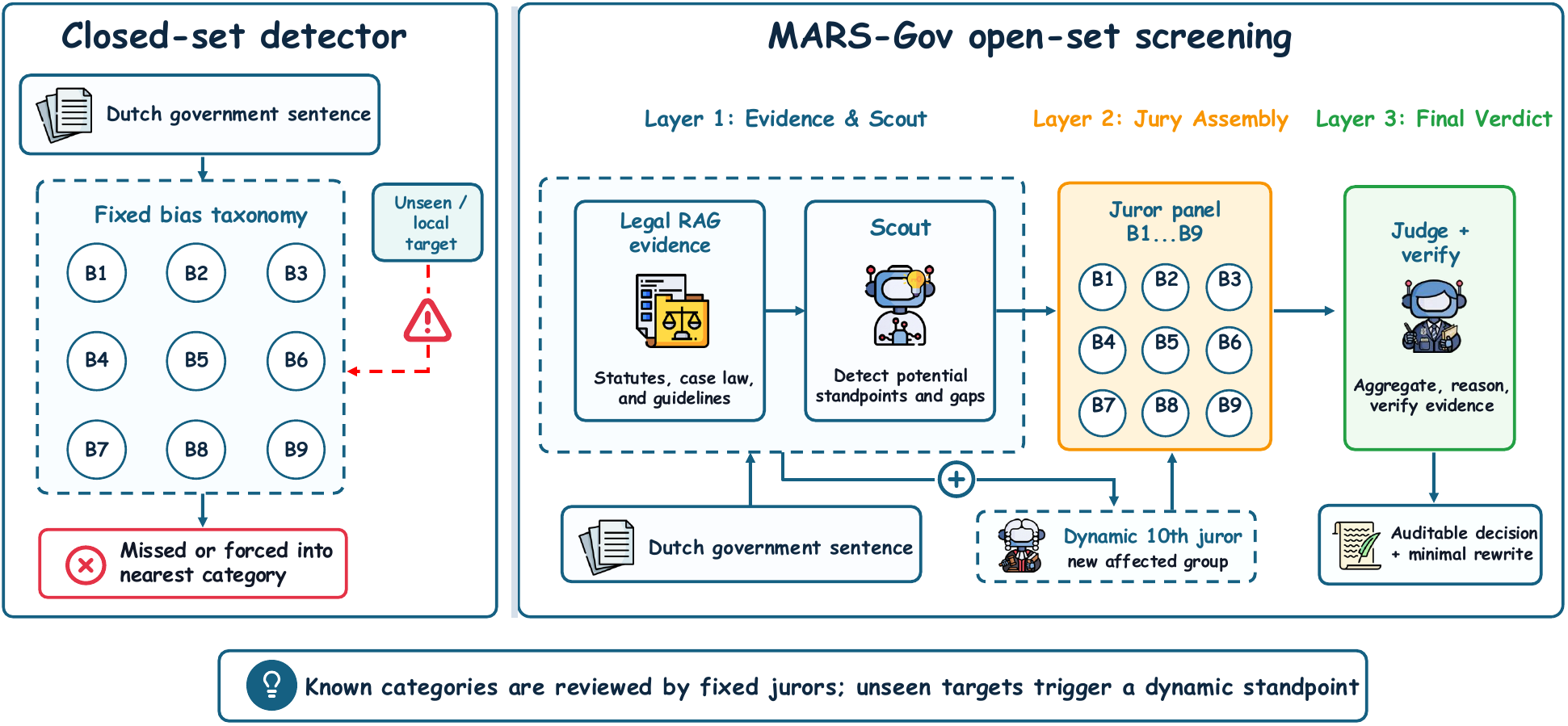}
    \caption{\textbf{Closed-set detection versus \model{} open-set screening.} Fixed taxonomies may miss local targets or force them into a known class. \model{} retrieves legal evidence, detects standpoint gaps, adds a dynamic 10th juror when needed, and sends grounded reports to the Judge for an auditable verdict and minimal rewrite.}
    \label{fig:closed_world}
\end{figure}

Existing approaches face three connected challenges. First, supervised encoders can be strong on closed benchmarks \citep{devlin-etal-2019-bert, chalkidis-etal-2020-legal}, but they mainly learn label regularities and provide limited evidence for why a decision is normatively justified \citep{blodgett-etal-2020-language}. Second, zero-shot LLMs are more flexible \citep{gallegos-etal-2024-bias}, but they often adopt a generic fairness viewpoint, over-flag group mentions, and give plausible verdicts without showing which legal or administrative norm supports the decision \citep{resnik-2025-llm-bias}. Third, both paradigms usually assume that the ontology of bias is fixed before evaluation begins.

A deeper problem is therefore the fixed label space. DGDB contains 3,747 Dutch government-document sentences with binary labels organized around nine bureaucratic-bias categories. As Figure~\ref{fig:closed_world} shows, a detector trained on these categories is evaluated as if all relevant categories were known in advance \citep{scheirer-etal-2013-openset}. Administrative language is less tidy. Digital illiteracy stigma, regional disadvantage, care dependency, or new socio-economic stereotypes may appear before an annotation scheme names them. Treating such cases as negative examples silently preserves the Closed-World Assumption; representing them only through the nearest fixed standpoint may preserve a binary alert while losing the affected group's specific rationale. A governance detector therefore needs both known-category review and open-set target discovery.

We propose \model{} (Multi-Agent Reasoning for Standpoint-aware Governance), a closed-loop framework for government text and a new SOTA system on DGDB. For each sentence, \model{} first retrieves legal and administrative context through Normative Evidence Contextualization (NEC). It then performs Open-Set Standpoint Screening (OSS): a Scout identifies potentially affected groups, fixed jurors inspect the nine known categories, and a dynamic ``10th juror'' is created when the Scout finds a target outside the panel. An Alert-based Routing Gate (ARG) conservatively clears low-risk sentences and escalates uncertain cases to a grounded Judge. If the Judge flags bias, a Restoration Agent proposes a minimal rewrite, and the system verifies the rewrite with the same detection criteria. The result is not a single classifier with a longer prompt, but an auditable decision loop for detection, mitigation, and review.

Our contributions are:
\begin{itemize}
    \setlength\itemsep{0em}
    \item We formulate bureaucratic bias detection as closed-loop governance: detection, evidence retrieval, minimal rewriting, and verification.
    \item We introduce \model{}, a legal-retrieval and juror-based framework that sets a new SOTA on DGDB.
    \item We show open-set and cross-lingual robustness through a dynamic ``10th juror'' and governance metrics for mitigation quality and over-governance.
\end{itemize}

\begin{figure*}[t]
  \centering
  \includegraphics[page=1,width=\textwidth]{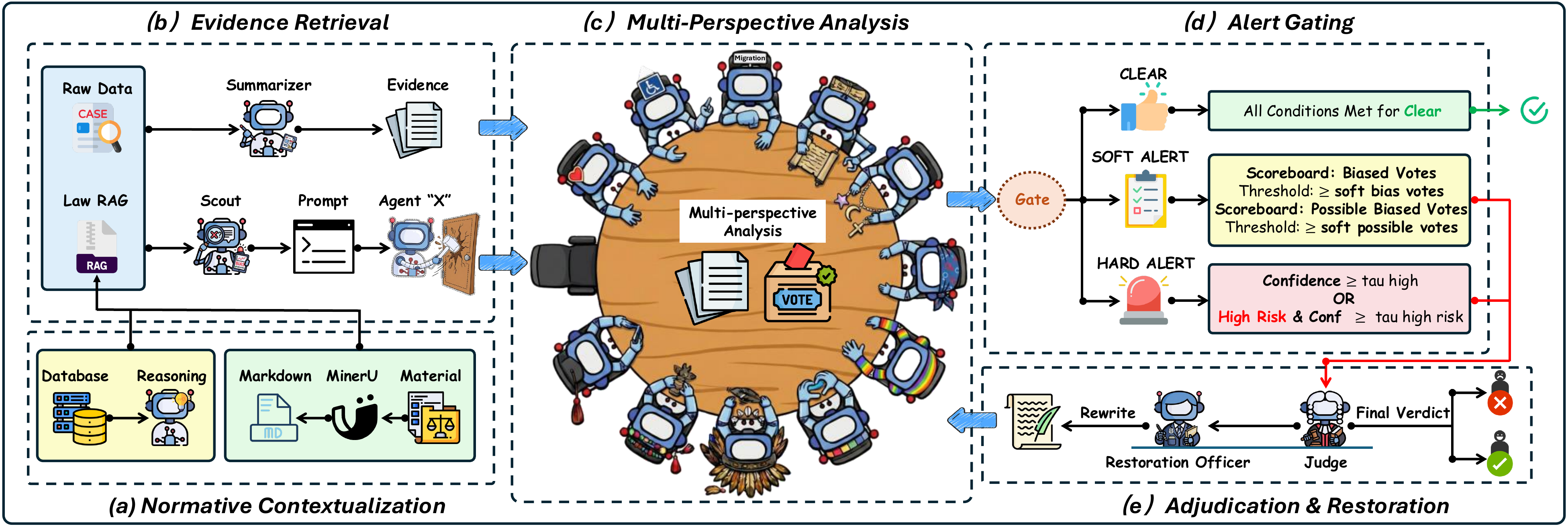}
  \caption{\textbf{Overview of \model{}.} The framework retrieves normative evidence, gathers juror risk reports, routes only risky cases to a Judge, and verifies any minimal rewrite. Separating evidence, routing, adjudication, and restoration keeps the decision trace inspectable.}
  \label{fig:overview}
\end{figure*}

\section{Related Work}

Bias in government documents is shaped by institutional practice as well as model behavior \citep{eubanks-2018-automating,noble-2018-algorithms,benjamin-2019-race,rizk-lindgren-2025-adm}. Legal NLP resources such as LEGAL-BERT, LexGLUE, and LegalBench-RAG support this setting \citep{chalkidis-etal-2020-legal,chalkidis-etal-2022-lexglue,chen-etal-2024-legalbenchrag}, while DGDB shows that bureaucratic bias often appears through framing and presupposition rather than slurs \citep{deswart-etal-2025-dgdb}. This view is consistent with discourse, framing, stereotyping, and intersectionality research, which treats bias as situated and socially produced \citep{fairclough-1992-discourse,vandijk-1993-principles,goffman-1974-frame,entman-1993-framing,otmakhova-etal-2024-media,allport-1954-prejudice,fiske-1998-stereotyping,crenshaw-1989-demarginalizing}. For administrative review, this matters because the same phrase can be harmless in a procedural context and discriminatory when it assigns risk, competence, or trustworthiness to a protected or stigmatized group.

\textbf{Supervised and benchmark-based bias detection.}
Computational bias detection has relied on association tests, debiasing methods, and BERT-style encoders \citep{caliskan-etal-2017-weat,bolukbasi-etal-2016-debias,devlin-etal-2019-bert}. Benchmarks such as StereoSet, BBQ, WinoBias, Social Bias Probing, and F2Bench probe stereotypes or fairness behavior with controlled examples \citep{nadeem-etal-2021-stereoset,parrish-etal-2022-bbq,zhao-etal-2018-winobias,marchiori-manerba-etal-2024-social,lan-etal-2025-f2bench}. These resources are useful, and fine-tuned encoders remain strong closed-set baselines, but their fixed labels or demographic contrasts make them brittle for newly emerging administrative targets. They also give little support for the later governance question: what should be changed, and how can the system know that the change did not distort the original administrative meaning?

\textbf{LLM-based detection, audits, and retrieval.}
LLMs can apply natural-language criteria without task-specific fine-tuning \citep{gallegos-etal-2024-bias,resnik-2025-llm-bias}. Audits nevertheless show hidden social associations, toxic continuations, long-form bias, and false positives around group mentions \citep{smith-etal-2022-im,blodgett-etal-2020-language,gehman-etal-2020-realtoxicityprompts,jeung-etal-2025-large,waseem-hovy-2016-hateful,dixon-etal-2018-unintended}. Retrieval can add legal evidence, but retrieval alone does not decide which standpoint should evaluate a sentence, how conflicting standpoints should be adjudicated, or whether a rewrite has repaired the problem. These unresolved steps are precisely where an apparently helpful LLM review can become hard to audit.

\textbf{Open-set and agentic governance.}
Open-set recognition and out-of-distribution detection study cases outside known categories \citep{scheirer-etal-2013-openset,hendrycks-gimpel-2017-baseline}; recent debiasing work applies related ideas to LMs \citep{wang-etal-2024-opendebias}. Multi-agent LLM systems coordinate specialized roles for software, simulation, safety review, and factuality \citep{wu-etal-2023-autogen,li-etal-2023-camel,qian-etal-2024-chatdev,hong-etal-2024-metagpt,park-etal-2023-generative,he2025aetheriamultimodalinterpretablecontent,ning2025madfactmultiagentdebateframework}. Most such systems define agent roles in advance. We combine legal grounding, open-set target discovery, standpoint-specialized agents, and closed-loop rewriting so that a newly affected group can receive a dedicated review standpoint instead of being represented only through the nearest fixed one.

\section{Methodology}
\label{sec:method}

\subsection{Task Definition: Closed-loop Bias Governance}
\label{sec:task}

We formulate \emph{closed-loop bias governance} for a sentence $x \in \mathcal{X}$ from a government document or benchmark. In DGDB, $x$ is a Dutch administrative sentence with an expert bias label. Unlike one-shot classification, the system may intervene and must verify the intervention. A pipeline $D(\cdot)$ returns an initial label $\hat{y}=D(x)\in\{0,1\}$, an optional minimal restoration $x'=R(x)$, and a detection-only verification $\hat{y}'=D_{\mathrm{verify}}(x')$ for non-empty restorations. The output also stores retrieved evidence and juror reports for audit. The task optimizes detection reliability, mitigation, and controlled side effects (\S\ref{sec:metrics}). This framing treats a biased sentence as a governance failure only if the system can identify the issue, propose a faithful repair, and confirm that the repair is no longer judged biased.

Two constraints follow from this formulation. First, the system must explain why a sentence is problematic in terms that can be checked against external norms, rather than only against the model's internal label boundary. Second, the rewrite must remain narrow: it should remove the biased implication without changing the administrative fact pattern. We therefore treat evidence retrieval, standpoint review, adjudication, restoration, and verification as separate artifacts in the trace.

\subsection{Overview: Pipeline-defined Detector}
\label{sec:overview}

Figure~\ref{fig:overview} shows \model{} as a pipeline-defined detector rather than a single classifier. This design keeps evidence, standpoint assignment, routing, and intervention separate enough to audit. It first builds a neutral evidence summary $E(x)=\mathrm{NEC}(x;\mathcal{K})$ from knowledge base $\mathcal{K}$, then uses a Scout and juror panel to produce reports $\mathcal{J}(x)=\mathrm{OSS}(x,E(x),\mathcal{G}(x))$. ARG clears low-risk cases and sends the rest to a Judge; if restoration is produced, $D_{\mathrm{verify}}$ applies the same detection criteria with rewriting disabled.

The trace records the retrieved evidence identifiers, the Scout's affected-group hypotheses, all juror reports, the routing action, the Judge rationale, the proposed restoration, and the verification result. These records are not presented as legal authority, but they make the model's path from sentence to intervention inspectable.

\begin{algorithm}[t]
\small
\SetAlgoLined
\DontPrintSemicolon
\caption{Inference in \model{}.}
\label{alg:inference}
\KwIn{Sentence $x$; knowledge base $\mathcal{K}$; fixed jurors $\mathcal{B}_{\mathrm{fix}}=\{B_1,\dots,B_9\}$.}
\KwOut{Detection $\hat{y}$; optional restoration $x'$; verification $\hat{y}'$; trace $\mathcal{T}$.}
$E \gets \mathrm{NEC}(x;\mathcal{K})$\;
$\mathcal{G} \gets \mathrm{Scout}(x,E)$\;
$\mathcal{U} \gets \mathrm{Uncovered}(\mathcal{G},\mathcal{B}_{\mathrm{fix}})$\;
$\mathcal{B}_{\mathrm{dyn}} \gets \{\mathrm{Instantiate}(g,x,E):g\in\mathcal{U}\}$\;
$\mathcal{J} \gets \mathrm{OSS}(x,E,\mathcal{B}_{\mathrm{fix}}\cup\mathcal{B}_{\mathrm{dyn}})$\tcp*{one report per fixed or dynamic juror}
$a \gets \mathrm{ARG}(\mathcal{G},\mathcal{J})$\tcp*{rule-based routing gate}
$\hat{y}\gets 0;\; x'\gets \emptyset;\; \hat{y}'\gets \emptyset$\;
\If{$a\neq\textsc{Clear}$}{
    $\hat{y}\gets \mathrm{Judge}(x,E,\mathcal{J})$\;
    \If{$\hat{y}=1$}{
        $x'\gets R(x,E,\mathcal{J})$\tcp*{minimal restoration}
        \If{$x'\neq\emptyset$}{
            $\hat{y}'\gets D_{\mathrm{verify}}(x')$\tcp*{same criteria; restoration disabled}
        }
    }
}
$\mathcal{T}\gets\{E,\mathcal{G},\mathcal{J},a,\hat{y},x',\hat{y}'\}$\;
\Return{$(\hat{y},x',\hat{y}',\mathcal{T})$}\;
\end{algorithm}

\subsection{Normative Evidence Contextualization via RAG (NEC)}
\label{sec:nec}

\begin{figure}[h]
    \centering
    \includegraphics[width=\linewidth]{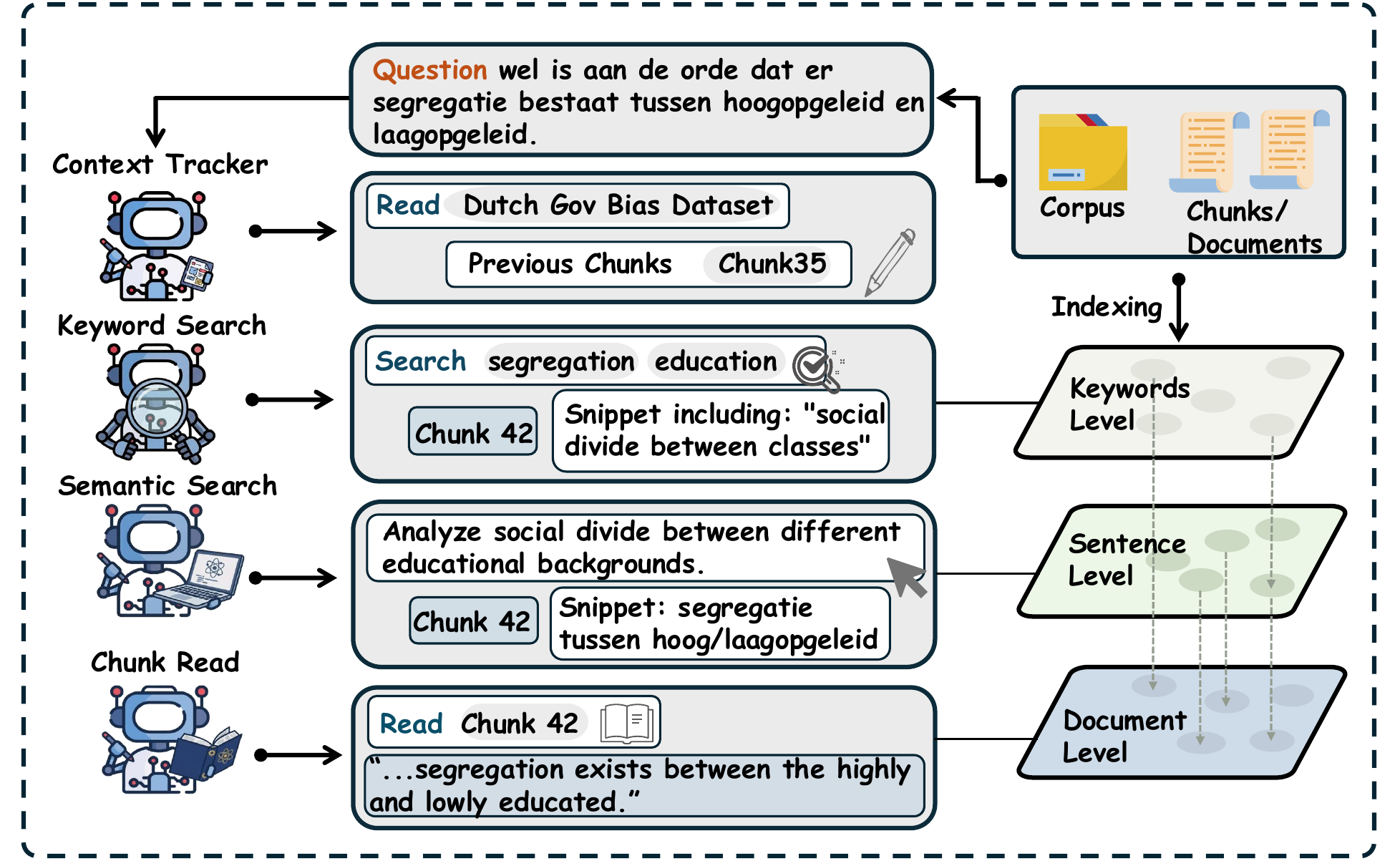}
    \caption{Three-layer retrieval in NEC. }
    \label{fig:arag}
\end{figure}

Governance decisions depend on policy definitions, protected attributes, and domain guidance, not only sentence semantics. NEC gives all downstream agents the same reference frame by applying three-stage RAG over $\mathcal{K}$ (Dutch equal-treatment law, antidiscrimination policy, accessibility guidance, online-harm and institutional-racism reports, inclusive-language guidance, and DGDB category notes): keyword retrieval, semantic retrieval, and chunk-level context expansion. Figure~\ref{fig:arag} summarizes these three retrieval stages. For input $x$, the retrieved evidence is
\begin{equation}
\label{eq:nec_retrieve}
C(x)=\mathrm{Read}\!\big(\mathrm{Ret}_{\mathrm{kw}}(x;\mathcal{K}),\,\mathrm{Ret}_{\mathrm{sem}}(x;\mathcal{K});\,\mathcal{K}\big).
\end{equation}
NEC then produces a neutral summary
\begin{equation}
\label{eq:nec_summarize}
E(x)=\mathrm{Summ}(x,C(x)).
\end{equation}
The summarizer returns definitions, criteria, examples, and short caveats, but no verdict. This keeps NEC from becoming an implicit classifier and leaves later judgments traceable to $C(x)$ rather than to unobserved model priors. In administrative text, this separation is important: the same word can be neutral in a rule description and harmful when attached to a group as a proxy for risk or competence.

\subsection{Open-Set Standpoint Screening (OSS): Scout + Juror Panel}
\label{sec:oss}

Bias governance is open-set because affected groups may not fit a fixed taxonomy. Given the input and evidence, the Scout identifies potentially implicated groups or sensitive attributes:
\begin{equation}
\label{eq:scout_groups}
\mathcal{G}(x) = \mathrm{Scout}(x, E(x)).
\end{equation}
Let $\mathcal{B}_{\mathrm{fix}}=\{B_1,\ldots,B_9\}$ denote the fixed panel, which covers Disability, Migration, Colonialism, Religion, Gender, LGBTQ+, Culture, Education/Class, and Welfare. The Scout maps every target in $\mathcal{G}(x)$ to this panel and collects the uncovered targets in
\begin{equation}
\label{eq:uncovered_groups}
\mathcal{U}(x)=\{g\in\mathcal{G}(x):\mathrm{Cover}(g,\mathcal{B}_{\mathrm{fix}})=0\}.
\end{equation}
For every $g\in\mathcal{U}(x)$, OSS instantiates one temporary, evidence-conditioned juror:
\begin{equation}
\label{eq:dynamic_jurors}
\begin{aligned}
\mathcal{B}_{\mathrm{dyn}}(x,E)&=\{B_g:g\in\mathcal{U}(x)\},\\
B_g&=\mathrm{Instantiate}(g,x,E).
\end{aligned}
\end{equation}
Thus, a sentence with several uncovered standpoints receives a separate dynamic report for each standpoint rather than collapsing them into one generic analysis. DGDB predominantly assigns one primary category per sentence, so the common case creates a single additional juror, informally called the ``10th juror''; this name is not an architectural limit. Dynamic jurors are not permanent ontology classes, but temporary review standpoints grounded in the current sentence and evidence.

Each fixed or dynamic juror returns an evidence-grounded report:
\begin{equation}
\label{eq:juror_reports}
\begin{aligned}
\mathcal{B}(x,E)&=\mathcal{B}_{\mathrm{fix}}\cup\mathcal{B}_{\mathrm{dyn}}(x,E),\\
\mathcal{J}(x,E)&=\{J_B:B\in\mathcal{B}(x,E)\},\\
J_B&=B(x,E).
\end{aligned}
\end{equation}
Here $J_B=(b_B,p_B,r_B,\rho_B)$ contains a binary verdict, confidence, severity, and a short rationale grounded in $E(x)$, optionally with supporting spans. These reports are passed to routing and restoration, giving the Judge several grounded views rather than one undifferentiated LLM answer.

This design also limits a common failure mode in open-set detection. The system may notice an unfamiliar affected group, but the final decision still passes through the same Judge, evidence, and verification steps as known categories. Open-set discovery therefore expands who can be reviewed, while the downstream standard for intervention remains unchanged.

\subsection{Alert-based Routing Gate (ARG): Risk-sensitive Escalation}
\label{sec:arg}

To reduce cost without lowering recall, \model{} uses an \emph{alert-based routing gate} (ARG)\footnote{The concrete thresholds are selected on the validation set and reported in Appendix~\ref{app:hyperparam}.}. ARG maps $(\mathcal{G},\mathcal{J})$ to \textsc{Clear} or \textsc{Escalate}, clearing only when all risk signals are weak and well-formed. It escalates under three conditions: \emph{hard} alerts, when Scout confidence exceeds $\tau_{\mathrm{high}}$ or a high-risk watchlist threshold; \emph{soft} alerts, when multiple jurors give weaker but consistent signals; and \emph{fail-safe} alerts, when any juror output is missing, malformed, or unparsable.

\begin{equation}
\mathrm{ARG}(\mathcal{G},\mathcal{J}) =
\begin{cases}
\textsc{Clear}, & \text{no alert fires},\\
\textsc{Escalate}, & \text{otherwise}.
\end{cases}
\end{equation}

Escalated cases go to the Judge and, when needed, restoration. ARG is rule-based, so routing changes do not introduce extra backbone calls.

\subsection{Decision and Restoration: Judge and Verify with Minimal Rewriting}
\label{sec:restoration}

For escalated cases, the Judge resolves juror conflicts using $x$, $E(x)$, and $\mathcal{J}(x,E(x))$. It grounds the final decision in explicit evidence and outputs a binary label
\begin{equation}
\label{eq:judge_decision}
\hat{y} \;=\; \mathrm{Judge}\!\left(x,\,E(x),\,\mathcal{J}(x,E(x))\right) \in \{0,1\}.
\end{equation}
The Judge sees raw reports and evidence, but not the ARG verdict, which keeps adjudication separate from threshold tuning. This matters in practice because conservative routing should affect cost and recall, not the legal rationale of the final verdict.

If $\hat{y}=1$, a Restoration Agent generates a minimally edited sentence
\begin{equation}
\label{eq:restoration}
x' \;=\; R(x, E(x), \mathcal{J}(x,E(x))).
\end{equation}
The restoration removes biased framing while preserving factual content, meaning, and administrative register. It minimizes surface edits, returns exactly one revised sentence, and emits an empty string if no safe rewrite is possible. Empty restorations are treated as failed mitigation. The restored sentence is then verified:
\begin{equation}
\label{eq:verification}
\hat{y}' \;=\; D_{\mathrm{verify}}(x').
\end{equation}
A governance action succeeds when $\hat{y}=1$, $x'\neq\varnothing$, and $\hat{y}'=0$, so the same detector clears the non-empty restoration. This prevents a rewrite from being counted as successful merely because it was produced.

We deliberately avoid free-form paraphrasing. In pilot runs, broad rewrites often improved tone while changing who did what, which is unacceptable for administrative records. The restoration agent is therefore instructed to preserve named entities, dates, obligations, eligibility criteria, and material facts unless the biased wording itself must be replaced.

We measure meaning preservation by the cosine similarity between sentence embeddings of the original and restored text:
\begin{equation}
\label{eq:semantic_fidelity}
\mathrm{SF}(x,x') \;=\; \cos\!\left(\phi(x),\,\phi(x')\right),
\end{equation}
where $\phi(\cdot)$ is a fixed text encoder.\footnote{We employ \texttt{clips/e5-base-trm-nl}; weights are available at \url{https://huggingface.co/clips/e5-base-trm-nl}.} If no restoration is produced, $\mathrm{SF}(x,x')=0$.

\subsection{Pipeline-aware Governance Metrics}
\label{sec:metrics}

F1 only assesses the initial decision on $x$. It does not show whether restoration resolves the problem or whether the system intervenes on neutral text. We therefore add metrics for the full detect--restore--verify loop. Given $\mathcal{D}=\{(x_i,y_i)\}_{i=1}^N$, let $\hat{y}_i=D(x_i)$, $x_i'$ be the restored sentence, $\hat{y}'_i=D_{\mathrm{verify}}(x_i')$ when $x_i'\neq\varnothing$, and $m_i=\mathbb{I}[x_i'\neq\varnothing]$. We also compute $\mathrm{SF}_i=\mathrm{SF}(x_i,x_i')$, with $\mathrm{SF}_i=0$ when no rewrite is produced. Bias Mitigation Rate (BMR) measures the share of correctly detected biased cases that are restored and cleared:
\begin{equation}
\label{eq:bmr}
\mathrm{BMR} \;=\;
\frac{\sum_i m_i\,\mathbb{I}[y_i=1 \wedge \hat{y}_i=1 \wedge \hat{y}'_i=0]}
{\sum_i \mathbb{I}[y_i=1 \wedge \hat{y}_i=1]}.
\end{equation}

Semantic Fidelity (SF) averages meaning preservation over successfully mitigated cases:
\begin{equation}
\label{eq:sf_mean}
\overline{\mathrm{SF}} \;=\;
\frac{\sum_i m_i\,\mathrm{SF}_i \cdot \mathbb{I}[y_i=1 \wedge \hat{y}_i=1 \wedge \hat{y}'_i=0]}
{\sum_i m_i\,\mathbb{I}[y_i=1 \wedge \hat{y}_i=1 \wedge \hat{y}'_i=0]}.
\end{equation}

Bias Governance Score (BGS) combines coverage, mitigation, and fidelity:
\begin{equation}
\label{eq:bgs}
\mathrm{BGS} \;=\; \mathrm{Recall}\times \mathrm{BMR}\times \overline{\mathrm{SF}}.
\end{equation}
Including $\mathrm{Recall}$ prevents a pipeline from scoring well by governing only easy positives. Over-Governance Rate (OGR) measures unnecessary intervention on non-biased inputs, equivalent to the false-positive rate on the initial detector:
\begin{equation}
\label{eq:ogr}
\mathrm{OGR} \;=\;
\frac{\sum_i \mathbb{I}[y_i=0 \wedge \hat{y}_i=1]}{\sum_i \mathbb{I}[y_i=0]}.
\end{equation}
Together, BMR, $\overline{\mathrm{SF}}$, BGS, and OGR evaluate mitigation quality and side effects beyond one-shot classification. We report these components alongside F1 because a model can look strong under ordinary classification while over-rewriting neutral text, or while producing rewrites that do not survive verification.

\section{Experiments}
\label{sec:experiments}

\noindent We organize the experiments around five questions:

\noindent\textbf{RQ1:} Does \model{} outperform fine-tuned, zero-shot, and agentic baselines?\par
\noindent\textbf{RQ2:} Can the Scout recover held-out bias categories outside the fixed juror panel?\par
\noindent\textbf{RQ3:} Which modules account for the gains in detection and governance quality?\par
\noindent\textbf{RQ4:} How much inference cost does ARG save without changing the detector?\par
\noindent\textbf{RQ5:} Does the review procedure remain useful across domains and languages?

\subsection{Experimental Setup}

\noindent\textbf{Datasets.}\footnote{Dataset statistics are summarized in Appendix~\ref{app:dataset_details}.} We use DGDB~\citep{deswart-etal-2025-dgdb}, a Dutch administrative benchmark with nine bias categories. To test vocabulary generalization, we evaluate on a held-out subset, DGDB-Unseen, containing new bias terms. We also use DALC~\citep{caselli2021dalc}, a Dutch social-media offensive-language set; SBIC~\citep{sap2020social}, an English social-bias inference corpus; and KoBBQ~\citep{jin2024kobbq}, a Korean BBQ-style bias benchmark.

\noindent\textbf{Baselines.}
We compare fine-tuned PLMs (\textit{BERTje}, \textit{RobBERT}), zero-shot LLMs (\textit{GPT-4o}, \textit{Grok-3}, \textit{Gemini 2.5 Pro}, \textit{DeepSeek-R1}), and agentic controls: \textit{SA-RAG} and \textit{Standard Debate}. Two matched Gemini 2.5 Pro controls test whether stronger prompting can explain the gains. A fixed 12-shot CoT prompt uses nine biased demonstrations spanning the DGDB categories and three hard non-biased demonstrations, all selected from non-test data with fixed IDs and order. A \textit{Single-Call Integrated Prompt} receives the same retrieved NEC evidence, task definition, and output requirements as \model{}, but performs affected-group identification, classification, rationale generation, minimal rewriting, and self-checking in one LLM call. For governance metrics, all generative controls use the same frozen $D_{\mathrm{verify}}$ and fixed SF encoder; the Single-Call self-check does not determine BMR or BGS.

\begin{table*}[t]
\centering
\footnotesize
\renewcommand{\arraystretch}{0.93}
\setlength{\tabcolsep}{2.2pt}
\resizebox{\textwidth}{!}{%
\begin{tabular}{@{}lccccccccc@{}}
\toprule
& \textbf{Prec.} & \textbf{Rec.} & \textbf{F1} & \textbf{F1$_{\text{OOD}}$} & \textbf{$\Delta$Drop} & \textbf{OGR$\downarrow$} & \textbf{BMR} & \textbf{SF} & \textbf{BGS} \\\midrule

\rowcolor{cat_ref}\multicolumn{10}{@{}l@{}}{\textit{Fine-tuned}} \\
BERTje \cite{devries2019bertje} & \cc 0.842\,\scriptsize{$\pm$.017} & \cb 0.784\,\scriptsize{$\pm$.021} & \cc 0.812\,\scriptsize{$\pm$.014} & \ca 0.554\,\scriptsize{$\pm$.032} & \rd -31.8\%\,\scriptsize{$\pm$1.62} & \ra 3.1\%\,\scriptsize{$\pm$0.51} & - & - & - \\
RobBERT \cite{delobelle2020robbert} & \cc 0.839\,\scriptsize{$\pm$.019} & \cb 0.784\,\scriptsize{$\pm$.018} & \cc 0.811\,\scriptsize{$\pm$.013} & \ca 0.499\,\scriptsize{$\pm$.034} & \rd -38.5\%\,\scriptsize{$\pm$1.94} & \ra 3.6\%\,\scriptsize{$\pm$0.62} & - & - & - \\\midrule

\rowcolor{cat_zero}\multicolumn{10}{@{}l@{}}{\textit{I. Prompted LLMs (Zero-/Few-shot)}} \\
GPT-4o & \ca 0.514\,\scriptsize{$\pm$.025} & \cc 0.882\,\scriptsize{$\pm$.021} & \ca 0.649\,\scriptsize{$\pm$.017} & \ca 0.621\,\scriptsize{$\pm$.024} & \ra -4.3\%\,\scriptsize{$\pm$1.07} & \rc 16.4\%\,\scriptsize{$\pm$0.83} & \cb 0.872\,\scriptsize{$\pm$.016} & \cb 0.821\,\scriptsize{$\pm$.021} & \ca 0.631\,\scriptsize{$\pm$.021} \\
Grok-3 & \ca 0.484\,\scriptsize{$\pm$.031} & \ce \textbf{0.920}\,\scriptsize{$\pm$.017} & \ca 0.634\,\scriptsize{$\pm$.019} & \ca 0.598\,\scriptsize{$\pm$.029} & \rb -5.7\%\,\scriptsize{$\pm$1.18} & \rd 19.1\%\,\scriptsize{$\pm$0.94} & \cb 0.864\,\scriptsize{$\pm$.019} & \cb 0.808\,\scriptsize{$\pm$.024} & \ca 0.642\,\scriptsize{$\pm$.022} \\
Gemini 2.5 Pro & \ca 0.524\,\scriptsize{$\pm$.024} & \cd 0.907\,\scriptsize{$\pm$.018} & \ca 0.664\,\scriptsize{$\pm$.016} & \ca 0.636\,\scriptsize{$\pm$.022} & \ra -4.2\%\,\scriptsize{$\pm$0.95} & \rc 15.2\%\,\scriptsize{$\pm$0.72} & \cb 0.881\,\scriptsize{$\pm$.014} & \cb 0.827\,\scriptsize{$\pm$.018} & \ca 0.661\,\scriptsize{$\pm$.020} \\
Gemini 2.5 Pro (12-shot CoT) & \cb 0.642\,\scriptsize{$\pm$.021} & \cc 0.883\,\scriptsize{$\pm$.019} & \cc 0.743\,\scriptsize{$\pm$.017} & \cc 0.716\,\scriptsize{$\pm$.021} & \ra -3.6\% & \rb 9.6\%\,\scriptsize{$\pm$0.61} & \cc 0.914\,\scriptsize{$\pm$.015} & \cc 0.876\,\scriptsize{$\pm$.018} & \cc 0.707\,\scriptsize{$\pm$.018} \\
DeepSeek-R1 & \ca 0.549\,\scriptsize{$\pm$.023} & \cd 0.887\,\scriptsize{$\pm$.019} & \cb 0.678\,\scriptsize{$\pm$.015} & \cb 0.652\,\scriptsize{$\pm$.021} & \ra -3.8\%\,\scriptsize{$\pm$0.83} & \rb 13.4\%\,\scriptsize{$\pm$0.65} & \cc 0.893\,\scriptsize{$\pm$.013} & \cc 0.851\,\scriptsize{$\pm$.016} & \cb 0.674\,\scriptsize{$\pm$.017} \\\midrule

\rowcolor{cat_debate}\multicolumn{10}{@{}l@{}}{\textit{II. Standard Debate (No RAG)}} \\
Standard Debate (GPT-4o) & \cb 0.587\,\scriptsize{$\pm$.026} & \cc 0.862\,\scriptsize{$\pm$.020} & \cb 0.698\,\scriptsize{$\pm$.018} & \cb 0.671\,\scriptsize{$\pm$.026} & \ra -3.9\%\,\scriptsize{$\pm$1.13} & \rb 12.1\%\,\scriptsize{$\pm$0.77} & \cc 0.902\,\scriptsize{$\pm$.017} & \cc 0.868\,\scriptsize{$\pm$.019} & \cb 0.674\,\scriptsize{$\pm$.020} \\
Standard Debate (Grok-3) & \ca 0.549\,\scriptsize{$\pm$.029} & \cd 0.906\,\scriptsize{$\pm$.019} & \cb 0.684\,\scriptsize{$\pm$.021} & \ca 0.651\,\scriptsize{$\pm$.031} & \rb -4.8\%\,\scriptsize{$\pm$1.43} & \rc 14.5\%\,\scriptsize{$\pm$0.87} & \cc 0.878\,\scriptsize{$\pm$.019} & \cc 0.851\,\scriptsize{$\pm$.022} & \cb 0.677\,\scriptsize{$\pm$.024} \\
Standard Debate (Gemini 2.5 Pro) & \cb 0.612\,\scriptsize{$\pm$.023} & \cd 0.886\,\scriptsize{$\pm$.019} & \cb 0.724\,\scriptsize{$\pm$.018} & \cb 0.696\,\scriptsize{$\pm$.024} & \rb -3.9\%\,\scriptsize{$\pm$0.96} & \rb 10.7\%\,\scriptsize{$\pm$0.69} & \cc 0.908\,\scriptsize{$\pm$.016} & \cc 0.884\,\scriptsize{$\pm$.018} & \cc 0.711\,\scriptsize{$\pm$.021} \\
Standard Debate (DeepSeek-R1) & \cb 0.634\,\scriptsize{$\pm$.022} & \cc 0.882\,\scriptsize{$\pm$.018} & \cc 0.738\,\scriptsize{$\pm$.017} & \cb 0.711\,\scriptsize{$\pm$.023} & \ra -3.7\%\,\scriptsize{$\pm$0.86} & \rb 10.2\%\,\scriptsize{$\pm$0.63} & \cd 0.921\,\scriptsize{$\pm$.014} & \cc 0.892\,\scriptsize{$\pm$.017} & \cc 0.724\,\scriptsize{$\pm$.018} \\\midrule

\rowcolor{cat_agent}\multicolumn{10}{@{}l@{}}{\textit{III. Integrated and Agentic Controls}} \\
SA-RAG (GPT-4o) & \cb 0.667\,\scriptsize{$\pm$.020} & \cc 0.852\,\scriptsize{$\pm$.019} & \cc 0.748\,\scriptsize{$\pm$.015} & \cc 0.723\,\scriptsize{$\pm$.021} & \ra -3.3\%\,\scriptsize{$\pm$0.71} & \rb 8.9\%\,\scriptsize{$\pm$0.59} & \cc 0.913\,\scriptsize{$\pm$.015} & \cc 0.876\,\scriptsize{$\pm$.017} & \cb 0.682\,\scriptsize{$\pm$.017} \\
SA-RAG (Grok-3) & \ca 0.619\,\scriptsize{$\pm$.024} & \cd 0.896\,\scriptsize{$\pm$.017} & \cc 0.732\,\scriptsize{$\pm$.017} & \cc 0.700\,\scriptsize{$\pm$.024} & \rb -4.4\%\,\scriptsize{$\pm$1.04} & \rb 11.1\%\,\scriptsize{$\pm$0.71} & \cc 0.889\,\scriptsize{$\pm$.018} & \cc 0.862\,\scriptsize{$\pm$.021} & \cb 0.687\,\scriptsize{$\pm$.019} \\
SA-RAG (Gemini 2.5 Pro) & \cb 0.687\,\scriptsize{$\pm$.019} & \cc 0.876\,\scriptsize{$\pm$.018} & \cc 0.770\,\scriptsize{$\pm$.014} & \cc 0.748\,\scriptsize{$\pm$.019} & \ra -2.9\%\,\scriptsize{$\pm$0.62} & \ra 7.9\%\,\scriptsize{$\pm$0.51} & \cd 0.918\,\scriptsize{$\pm$.013} & \cc 0.893\,\scriptsize{$\pm$.014} & \cc 0.718\,\scriptsize{$\pm$.015} \\
Single-Call Integrated Prompt & \cd 0.762\,\scriptsize{$\pm$.019} & \cc 0.869\,\scriptsize{$\pm$.018} & \cd 0.812\,\scriptsize{$\pm$.015} & -- & -- & \ra 6.7\%\,\scriptsize{$\pm$0.55} & \cd 0.927\,\scriptsize{$\pm$.015} & \cd 0.912\,\scriptsize{$\pm$.017} & \cd 0.735\,\scriptsize{$\pm$.019} \\
SA-RAG (DeepSeek-R1) & \cc 0.710\,\scriptsize{$\pm$.018} & \cc 0.876\,\scriptsize{$\pm$.017} & \cc 0.784\,\scriptsize{$\pm$.013} & \cc 0.762\,\scriptsize{$\pm$.018} & \ra -2.8\%\,\scriptsize{$\pm$0.53} & \ra 7.2\%\,\scriptsize{$\pm$0.47} & \cd 0.929\,\scriptsize{$\pm$.012} & \cd 0.902\,\scriptsize{$\pm$.013} & \cc 0.734\,\scriptsize{$\pm$.013} \\\midrule

\rowcolor{cat_ours}\multicolumn{10}{@{}l@{}}{\textit{IV. Ours: \model{} Framework}} \\
\model{} (GPT-4o) & \cc 0.844\,\scriptsize{$\pm$.017} & \cc 0.863\,\scriptsize{$\pm$.015} & \cd 0.853\,\scriptsize{$\pm$.013} & \cd 0.823\,\scriptsize{$\pm$.019} & \ra -3.5\%\,\scriptsize{$\pm$0.62} & \ra 3.5\%\,\scriptsize{$\pm$0.32} & \cd 0.934\,\scriptsize{$\pm$.011} & \cd 0.939\,\scriptsize{$\pm$.013} & \cd 0.757\,\scriptsize{$\pm$.014} \\
\model{} (Grok-3) & \cc 0.836\,\scriptsize{$\pm$.019} & \ce \textbf{0.910}\,\scriptsize{$\pm$.013} & \cd 0.871\,\scriptsize{$\pm$.014} & \cd 0.840\,\scriptsize{$\pm$.022} & \ra -3.6\%\,\scriptsize{$\pm$0.77} & \ra 3.9\%\,\scriptsize{$\pm$0.41} & \cd 0.922\,\scriptsize{$\pm$.013} & \cd 0.928\,\scriptsize{$\pm$.015} & \cd 0.779\,\scriptsize{$\pm$.018} \\
\textbf{\model{} (Gemini 2.5 Pro)} & \cd 0.865\,\scriptsize{$\pm$.015} & \cd 0.895\,\scriptsize{$\pm$.013} & \ce \textbf{0.880}\,\scriptsize{$\pm$.012} & \ce \textbf{0.853}\,\scriptsize{$\pm$.017} & \ra -3.1\%\,\scriptsize{$\pm$0.51} & \ra \textbf{2.5\%}\,\scriptsize{$\pm$0.28} & \ce \textbf{0.951}\,\scriptsize{$\pm$.009} & \ce \textbf{0.949}\,\scriptsize{$\pm$.011} & \ce \textbf{0.808}\,\scriptsize{$\pm$.013} \\
\textbf{\model{} (DeepSeek-R1)} & \ce \textbf{0.872}\,\scriptsize{$\pm$.016} & \cd 0.884\,\scriptsize{$\pm$.014} & \cd 0.878\,\scriptsize{$\pm$.013} & \cd 0.857\,\scriptsize{$\pm$.017} & \ra \textbf{-2.4\%}\,\scriptsize{$\pm$0.47} & \ra 2.6\%\,\scriptsize{$\pm$0.29} & \ce \textbf{0.948}\,\scriptsize{$\pm$.011} & \ce 0.962\,\scriptsize{$\pm$.010} & \ce 0.806\,\scriptsize{$\pm$.015} \\\bottomrule

\end{tabular}%
}
\caption{\textbf{Main DGDB results.} Darker \colorbox{hg2}{green} marks stronger performance; darker \colorbox{hr2}{red} marks larger $\Delta$Drop or OGR. F1$_{\text{OOD}}$ is DGDB-Unseen F1; $\Delta$Drop is the relative drop from DGDB F1. Single-Call was evaluated in-domain only. \model{} gives the best mean F1/BGS with low governance intrusion.}
\label{tab:main_results_heatmap}
\vspace{-0.8em}
\end{table*}

\subsection{Overall Performance (RQ1)}
Table \ref{tab:main_results_heatmap} reports in-domain DGDB and DGDB-Unseen performance across four backbones\footnote{Additional experiments with more backbone models are reported in Appendix~\ref{app:extended_benchmarks}.}. DGDB-Unseen replaces selected DGDB bias terms with alternative unseen expressions while preserving labels and administrative context, testing whether detectors generalize beyond the original lexical cues.

\vspace{0.2em}
\noindent\ding{182} \textbf{Architecture Trumps Generic Debate.}
Zero-shot models have high recall but plateau at F1 $\approx 0.63\text{--}0.68$ because precision stays below 0.55. Standard Debate improves precision only modestly, and SA-RAG does not consistently surpass the best fine-tuned encoder. \model{} reaches SOTA F1 ($>0.85$) across all backbones, with Gemini 2.5 Pro obtaining the best mean F1 (0.880).

\noindent\textbf{Matched prompting controls.} The 12-shot CoT prompt raises Gemini 2.5 Pro F1 from 0.664 to 0.743 and lowers OGR from 15.2\% to 9.6\%. Integrating all steps into one call further raises F1 to 0.812 and BGS to 0.735. The full staged procedure remains 6.8 F1 points and 7.3 BGS points higher than Single-Call, while reducing OGR from 6.7\% to 2.5\%. These comparisons isolate gains beyond demonstrations, prompt length, and retrieved context alone.

\vspace{0.2em}
\noindent\ding{183} \textbf{DGDB-Unseen Exposes Lexical Fragility.}
Fine-tuned encoders look attractive on the original split: BERTje and RobBERT are compact and reach F1 $\approx 0.81$, higher than most Standard Debate variants. Yet their F1$_{\text{OOD}}$ falls to 0.554 and 0.499 ($\Delta$Drop = -31.8\% and -38.5\%). This is a serious weakness for bias detection, because social naming practices change and new stigmatizing terms appear after training. \model{} is much more stable, keeping DGDB-Unseen F1 at 0.823--0.857 with $\Delta$Drop between -3.6\% and -2.4\%.

\vspace{0.2em}
\noindent\ding{184} \textbf{Improving the Precision-Recall Balance.}
Grok-3 gives high zero-shot recall (0.920) but high OGR (19.1\%), showing that recall alone can produce excessive intervention. \model{} (Gemini 2.5 Pro) keeps recall high (0.895), raises precision to 0.865, and lowers OGR to 2.5\%. DeepSeek-R1 gives the highest \model{} precision (0.872) and the lowest DGDB-Unseen drop (-2.4\%).

Overall, the main gain comes from routing legal evidence and standpoint-specific reports into adjudication. Every backbone benefits more from \model{} than from Standard Debate or SA-RAG, and the best configuration improves BGS to 0.808 while keeping unnecessary intervention low.

\subsection{Controlled Open-set Evaluation with Held-out Categories (RQ2)}\label{sec:loco_evaluation}
\noindent\textbf{Open-set evaluation.} DGDB-Unseen tests lexical robustness: whether a detector remains stable when familiar bias terms are replaced by unseen expressions. It does not show whether \model{} can recover a bias category that is absent from the fixed juror panel. To test this stronger open-set claim, we design a leave-one-juror-out experiment over all 9 categories with Gemini 2.5 Pro; the setup is as follows.

\noindent\textbf{Design.} Each DGDB category is associated with a fixed juror and watchlist. We remove one juror--watchlist pair at a time and evaluate the corresponding test subset, excluding both the specialist and its lexical cues. We encode the Scout's predicted group and rationale together with fixed category prototypes using \texttt{clips/e5-base-trm-nl}, and rank the categories by cosine similarity. Correct@$k$ indicates whether the held-out category appears among the top $k$ predictions. The prototypes are fixed before evaluation and are not tuned by category; Appendix~\ref{app:loco_texts} provides the full texts.

\begin{table*}[t]
    \centering
    \footnotesize
    \renewcommand{\arraystretch}{0.92}
    \resizebox{\textwidth}{!}{%
    \begin{tabular}{llccccccccc}
    \toprule
    \textbf{Config.} & \textbf{Metric} & \textbf{B1} & \textbf{B2} & \textbf{B3} & \textbf{B4} & \textbf{B5} & \textbf{B6} & \textbf{B7} & \textbf{B8} & \textbf{B9} \\
    & & \scriptsize (Disability) & \scriptsize (Migration) & \scriptsize (Colonialism) & \scriptsize (Religion) & \scriptsize (Gender) & \scriptsize (LGBTQ+) & \scriptsize (Culture) & \scriptsize (Education/Class) & \scriptsize (Welfare) \\
    \midrule
    \multirow{2}{*}{w/o Scout}
    & Rec.             & 0.794\,\std{.027} & 0.814\,\std{.024} & 0.778\,\std{.031} & 0.776\,\std{.028} & 0.821\,\std{.023} & 0.788\,\std{.026} & 0.807\,\std{.024} & 0.764\,\std{.029} & 0.812\,\std{.022} \\
    & F1               & 0.804\,\std{.021} & 0.826\,\std{.018} & 0.794\,\std{.025} & 0.789\,\std{.022} & 0.831\,\std{.017} & 0.806\,\std{.019} & 0.813\,\std{.018} & 0.782\,\std{.023} & 0.819\,\std{.017} \\
    \midrule
    \multirow{4}{*}{\shortstack[l]{\textbf{w/ Scout} \\ \textbf{(Ours)}}}
    & Rec.             & 0.882\,\std{.018} & 0.901\,\std{.015} & 0.874\,\std{.021} & 0.871\,\std{.019} & 0.894\,\std{.014} & 0.881\,\std{.018} & 0.889\,\std{.016} & 0.866\,\std{.021} & 0.896\,\std{.015} \\
    & \textbf{F1}      & \textbf{0.873}\,\std{.015} & \textbf{0.891}\,\std{.012} & \textbf{0.867}\,\std{.017} & \textbf{0.864}\,\std{.015} & \textbf{0.886}\,\std{.011} & \textbf{0.872}\,\std{.014} & \textbf{0.881}\,\std{.013} & \textbf{0.854}\,\std{.017} & \textbf{0.884}\,\std{.012} \\
    \cmidrule{2-11}
    & Correct@1 (\%)   & 83.4\,\std{2.4} & 87.6\,\std{1.9} & 82.7\,\std{2.7} & 86.5\,\std{2.1} & 88.9\,\std{1.7} & 84.3\,\std{2.2} & 81.6\,\std{2.6} & 83.2\,\std{2.5} & 87.4\,\std{1.8} \\
    & Correct@3 (\%)   & 92.5\,\std{1.5} & 96.3\,\std{0.9} & 91.8\,\std{1.7} & 94.7\,\std{1.2} & 95.6\,\std{1.1} & 93.4\,\std{1.4} & 92.7\,\std{1.6} & 92.1\,\std{1.6} & 95.2\,\std{1.1} \\
    \bottomrule
    \end{tabular}%
    }
    \caption{Leave-One-Category-Out (LOCO) evaluation on Gemini 2.5 Pro. Best F1 scores are bold. \textit{Significance:} all w/ Scout gains over w/o Scout remain significant after Bonferroni correction over 9 categories (two-sided paired t-test over 3 runs; max corrected $p=2.4{\times}10^{-2}$).}
    \label{tab:loco_evaluation}
\end{table*}

\noindent\textbf{Results.} Table~\ref{tab:loco_evaluation} shows that, without the Scout, the remaining jurors achieve an average F1 of 0.807: they often detect broad negative framing but fail to identify the held-out category. Scout-guided dynamic review raises average F1 to 0.875 and recovers the held-out category with 85.1\% Correct@1 and 93.8\% Correct@3. The gains are pronounced for categories with indirect cues, including Colonialism and Education/Class. This distinction matters for rewriting: a neighboring juror may flag the sentence as biased, but recovering the held-out standpoint provides the category-specific basis for revision.

\subsection{Ablation \& Efficiency Analysis (RQ3--RQ4)}\label{sec:rq5_ablation}

\noindent\textbf{Detection modules.} Table \ref{tab:ablation} shows that Scout-guided dynamic juror review prevents static-juror consensus errors (F1 $\downarrow$ 0.035 without it and recall $\downarrow$ 5.5\%). Removing NEC reduces F1 to 0.824, showing that legal grounding helps separate neutral bureaucratic phrasing from normative violation, especially in cases that are lexically neutral but procedurally stigmatizing.

\begin{table}[t]
\centering
\resizebox{\columnwidth}{!}{%
\renewcommand{\arraystretch}{1.0}
\setlength{\tabcolsep}{2.5pt}
\begin{tabular}{l|ccc|c|c}
\toprule
\multirow{2}{*}{\textbf{Configuration}}
& \multicolumn{3}{c|}{\textbf{Detection Quality}}
& \textbf{Governance}
& \textbf{Efficiency} \\
& \textbf{Prec.}
& \textbf{Rec.}
& \textbf{F1}
& \textbf{BGS}
& \textbf{Tokens/Doc} \\
\midrule
\rowcolor{gray!10}
\textbf{Full \model{}}
& \textbf{0.865}\,\scriptsize{$\pm$.015}
& \textbf{0.895}\,\scriptsize{$\pm$.013}
& \textbf{0.880}\,\scriptsize{$\pm$.012}
& \textbf{0.808}\,\scriptsize{$\pm$.013}
& \textbf{2,450}\,\scriptsize{$\pm$58} \\
\midrule
\textit{w/o Verify Loop}
& 0.865\,\scriptsize{$\pm$.015}
& 0.895\,\scriptsize{$\pm$.013}
& 0.880\,\scriptsize{$\pm$.012}
& 0.646\,\scriptsize{$\pm$.019}
  \textcolor{red}{($\downarrow$ 20.0\%)$^{\ast\ast}$}
& 2,100\,\scriptsize{$\pm$49} \\
\textit{w/o Reasoning}
& 0.842\,\scriptsize{$\pm$.019}
& 0.870\,\scriptsize{$\pm$.017}
& 0.856\,\scriptsize{$\pm$.016}
& 0.765\,\scriptsize{$\pm$.021}
  ($\downarrow$ 5.3\%)$^{\ast}$
& 2,380\,\scriptsize{$\pm$62} \\
\textit{w/o NEC}
& 0.802\,\scriptsize{$\pm$.021}
& 0.847\,\scriptsize{$\pm$.019}
& 0.824\,\scriptsize{$\pm$.018}
& 0.710\,\scriptsize{$\pm$.022}
  ($\downarrow$ 12.1\%)$^{\ast\ast}$
& 2,250\,\scriptsize{$\pm$54} \\
\textit{w/o Scout (10th)}
& 0.851\,\scriptsize{$\pm$.017}
& 0.840\,\scriptsize{$\pm$.019}
& 0.845\,\scriptsize{$\pm$.015}
& 0.760\,\scriptsize{$\pm$.018}
  ($\downarrow$ 5.9\%)$^{\ast}$
& 2,150\,\scriptsize{$\pm$44} \\
\midrule
\textit{w/o ARG}
& 0.865\,\scriptsize{$\pm$.015}
& 0.895\,\scriptsize{$\pm$.013}
& 0.880\,\scriptsize{$\pm$.012}
& 0.808\,\scriptsize{$\pm$.013}
& 6,100\,\scriptsize{$\pm$131}
  \textcolor{red}{($\uparrow$ 149\%)} \\
\bottomrule
\end{tabular}%
}
\caption{Ablation on DGDB with Gemini 2.5 Pro, averaged over 3 runs. \textit{Significance:} marked BGS drops vs. Full \model{} remain significant after Bonferroni correction (two-sided paired t-test; $^{\ast\ast}p<0.01$, $^{\ast}p<0.05$; max corrected $p=2.7{\times}10^{-2}$). w/o ARG is cost-only.}
\label{tab:ablation}
\end{table}

\begin{figure}[t]
    \centering
    \hspace*{-0.5em}% shift slightly left to keep the rightmost tick clear
    \includegraphics[width=0.84\columnwidth]{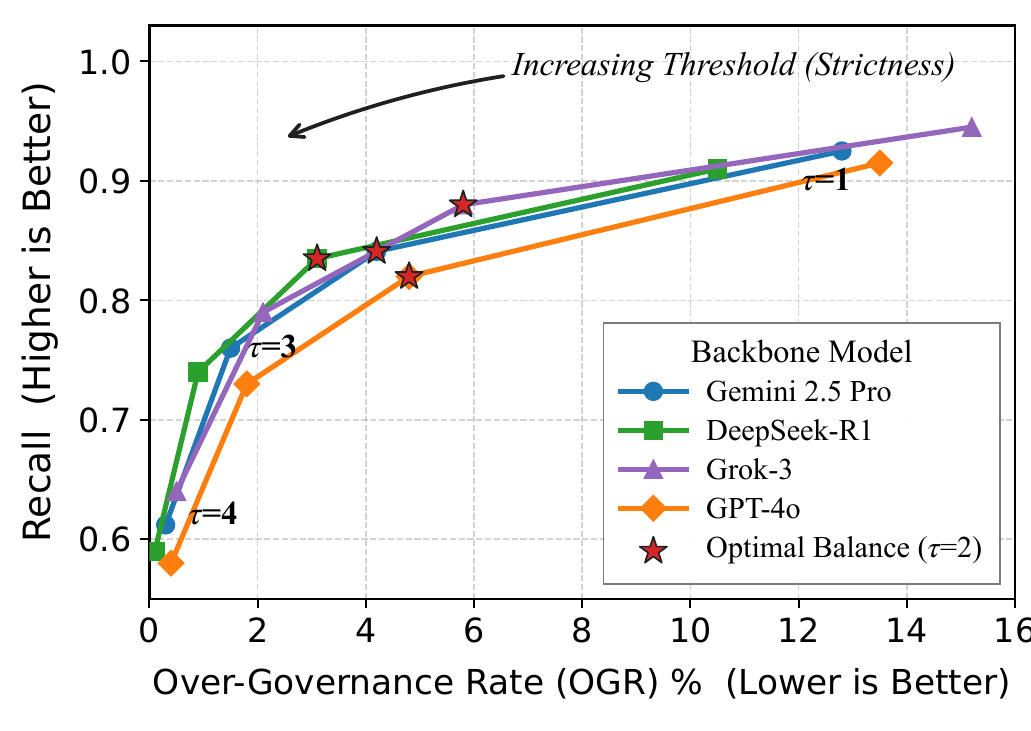}
    \caption{Recall vs. Over-Governance Rate (OGR) trade-off as threshold $\tau$ increases.}
    \label{fig:tradeoff_multi}
\end{figure}

\noindent\textbf{Governance \& efficiency.} Removing verification drops BGS from 0.808 to 0.646 and inflates BMR to 98\%, because rewrites are no longer checked against the same detector.\footnote{The ARG hyperparameter grid search is in Appendix~\ref{app:hyperparam}.} ARG filters about 65\% of documents as \textsc{Clear}, reducing token use by roughly 60\%. Figure~\ref{fig:tradeoff_multi} shows a stable elbow at $\tau=2$, where over-governance falls with little recall loss across backbones.

Table~\ref{tab:efficiency_cost} reports cost. The first-pass detector uses NEC, the Scout, and nine fixed jurors; dynamic jurors are called only for uncovered targets, one per target, and the rule-based gate adds no LLM call. With Gemini 2.5 Pro, ARG clears 64.8\% of inputs early and reduces calls from 28.6 to 12.4, latency from 41.5s to 19.2s, and cost from \$18.30 to \$7.35 per 1,000 sentences. Since removing the gate leaves detection unchanged but roughly doubles cost, ARG acts as an efficiency component rather than a hidden classifier.

\begin{table}[t]
\centering
\footnotesize
\setlength{\tabcolsep}{2.5pt}
\renewcommand{\arraystretch}{1.0}
\resizebox{\columnwidth}{!}{%
\begin{tabular}{l c c c c}
\toprule
\textbf{Configuration}
& \textbf{Calls}
& \textbf{Lat.\,(s)}
& \textbf{Clear\%}
& \textbf{\$/1k} \\
\midrule
\rowcolor{cat_zero}
\multicolumn{5}{@{}l@{}}{\textit{Baselines (Gemini 2.5 Pro)}} \\
Zero-shot LLM
& 1.0
& 1.8 (0.3)
& ---
& 0.85 \\
Standard Debate
& 4.0
& 7.2 (0.9)
& ---
& 4.62 \\
SA-RAG
& 1.0
& 3.4 (0.5)
& ---
& 2.46 \\
\midrule
\rowcolor{cat_ours}
\multicolumn{5}{@{}l@{}}{\textit{\model{} (with ARG)}} \\
GPT-4o
& 12.6 (0.6)
& 20.5 (2.3)
& 64.2 (1.7)
& 10.04 \\
Grok-3
& 12.8 (0.7)
& 18.7 (2.1)
& 63.5 (1.9)
& 13.39 \\
\textbf{Gemini 2.5 Pro}
& \textbf{12.4 (0.5)}
& \textbf{19.2 (2.2)}
& \textbf{64.8 (1.6)}
& \textbf{7.35} \\
DeepSeek-R1
& 12.5 (0.5)
& 31.6 (3.2)
& 65.1 (1.8)
& 4.25 \\
\midrule
\rowcolor{cat_ref}
\textit{w/o ARG (Gemini)}
& 28.6 (0.8)
& 41.5 (3.7)
& ---
& 18.30 \\
\bottomrule
\end{tabular}
}
\caption{Per-sentence inference cost on DGDB (1{,}000 sentences, 3 runs; mean with $\sigma$). \emph{Calls} counts all backbone calls; \emph{Clear\%} is the early-stop share.}
\label{tab:efficiency_cost}
\end{table}

\begin{table*}[t]
\centering
\footnotesize
\resizebox{\textwidth}{!}{%
\begin{tabular}{ l | ccc | ccc | ccc }
\toprule
\multirow{2}{*}{\textbf{Configuration}} & \multicolumn{3}{c|}{\textbf{DALC} (Dutch, Social Media)} & \multicolumn{3}{c|}{\textbf{SBIC} (English, Social Media)} & \multicolumn{3}{c}{\textbf{KoBBQ} (Korean, General)} \\
& \textbf{Prec.} & \textbf{Rec.} & \textbf{F1} & \textbf{Prec.} & \textbf{Rec.} & \textbf{F1} & \textbf{Prec.} & \textbf{Rec.} & \textbf{F1} \\
\midrule
Zero-shot LLMs & 0.513\,\scriptsize{$\pm$.028} & 0.873\,\scriptsize{$\pm$.021} & 0.646\,\scriptsize{$\pm$.018} & 0.486\,\scriptsize{$\pm$.033} & 0.844\,\scriptsize{$\pm$.023} & 0.617\,\scriptsize{$\pm$.019} & 0.451\,\scriptsize{$\pm$.036} & 0.803\,\scriptsize{$\pm$.027} & 0.578\,\scriptsize{$\pm$.024} \\
Standard Debate & 0.607\,\scriptsize{$\pm$.025} & 0.851\,\scriptsize{$\pm$.020} & 0.709\,\scriptsize{$\pm$.019} & 0.583\,\scriptsize{$\pm$.027} & 0.819\,\scriptsize{$\pm$.021} & 0.681\,\scriptsize{$\pm$.021} & 0.553\,\scriptsize{$\pm$.029} & 0.811\,\scriptsize{$\pm$.024} & 0.658\,\scriptsize{$\pm$.023} \\
SA-RAG & 0.653\,\scriptsize{$\pm$.021} & 0.862\,\scriptsize{$\pm$.018} & 0.743\,\scriptsize{$\pm$.016} & 0.626\,\scriptsize{$\pm$.023} & 0.836\,\scriptsize{$\pm$.019} & 0.716\,\scriptsize{$\pm$.018} & 0.593\,\scriptsize{$\pm$.027} & 0.821\,\scriptsize{$\pm$.021} & 0.689\,\scriptsize{$\pm$.019} \\
\rowcolor{cat_ours} \textbf{Ours: \model{}} & \textbf{0.823}\,\scriptsize{$\pm$.015} & \textbf{0.886}\,\scriptsize{$\pm$.013} & \textbf{0.853}\,\scriptsize{$\pm$.012} & \textbf{0.803}\,\scriptsize{$\pm$.018} & \textbf{0.847}\,\scriptsize{$\pm$.016} & \textbf{0.824}\,\scriptsize{$\pm$.015} & \textbf{0.776}\,\scriptsize{$\pm$.021} & \textbf{0.829}\,\scriptsize{$\pm$.018} & \textbf{0.802}\,\scriptsize{$\pm$.017} \\
\bottomrule
\end{tabular}%
}
\caption{Generalization on DALC, SBIC, and KoBBQ using Gemini 2.5 Pro. Results average 3 runs; best F1 scores are bold. \textit{Significance:} \model{} beats SA-RAG after Bonferroni correction over 3 datasets (two-sided paired t-test; max corrected $p=1.1{\times}10^{-2}$).}
\label{tab:external_generalization}
\end{table*}

\subsection{Cross-Domain and Cross-Lingual Generalization (RQ5)}
\label{sec:generalization_external}
We test transfer on DALC, SBIC, and KoBBQ using balanced 1,000-instance samples and Gemini 2.5 Pro. Table~\ref{tab:external_generalization} shows that \model{} remains strongest across all three datasets. The gain mainly comes from precision: the review loop rejects unsupported group-risk inferences left by zero-shot detection, adding 20--30 precision points while keeping recall comparable. Absolute F1 remains below in-domain DGDB because the external benchmarks shift genre, label assumptions, and language.

These results support the portability of the review procedure, not of Dutch law itself: deployment in another jurisdiction would still need a local legal and policy evidence base.

\section{Conclusion}
We presented \model{}, a standpoint-aware framework for closed-loop bias governance in Dutch government documents. Rather than treating bias detection as one-shot classification over a fixed taxonomy, \model{} combines legal retrieval, open-set target screening, juror-based adjudication, conservative routing, and rewrite verification. Experiments show stronger performance than fine-tuned, zero-shot, and generic debate baselines, with higher open-set recovery and BGS, better cross-domain and cross-lingual transfer, and lower over-governance. These results suggest that reliable governance NLP requires more than stronger classifiers: it requires legally grounded evidence, explicit standpoint reasoning, and verification after interventions. The DGDB-Unseen results sharpen this point: compact encoders remain competitive on familiar terms, but lose reliability when public language shifts. Future work will extend the framework to additional jurisdictions and reduce inference cost for broader evaluation and lower-cost deployment.

\section*{Limitations}
Our framework remains dependent on domain-specific normative resources. Although the external experiments suggest that the core mechanism can transfer across languages and domains, \model{}'s strongest setting is Dutch administrative text because its evidence base is built around Dutch anti-discrimination law and administrative guidance. Applying the framework to other jurisdictions or platforms therefore requires localized legal and policy resources, not prompt translation alone; otherwise, the system may produce justifications that are linguistically plausible but normatively incorrect.

A second limitation is inference cost. The multi-agent pipeline is more expensive and slower than lightweight discriminative models, even with the routing gate. This cost is acceptable for high-stakes public-sector review, where false positives and false negatives can both have institutional consequences, but it limits use in high-throughput or real-time settings. Reducing the number of model calls without weakening legal grounding or verification remains an important direction for future work.

Finally, BGS is an aggregate governance metric rather than a complete diagnostic. Because $\mathrm{BGS} = \mathrm{Recall} \times \mathrm{BMR} \times \overline{\mathrm{SF}}$ (Eq.~\ref{eq:bgs}), different systems can obtain similar scores through different trade-offs between detection coverage, mitigation success, and meaning preservation. We therefore report the individual components alongside BGS in Table~\ref{tab:main_results_heatmap}. BGS should be interpreted as a compact summary of governance performance, not as a substitute for component-level analysis.

\section*{Ethics Statement}
\label{sec:ethics}

This work addresses a socially beneficial but high-stakes NLP setting: identifying potentially biased language in Dutch government documents and proposing minimally edited alternatives. \model{} is not intended to act as an autonomous decision-maker or to automatically rewrite official records. In any real deployment, it should be used only as a human-in-the-loop decision-support system: it may flag potentially problematic passages, present evidence and rationales, and suggest candidate rewrites for review by qualified experts such as legal professionals, policy officers, or ethics reviewers. The original text must remain preserved in the auditable record, especially in historical or contested cases, so that affected individuals retain access to recourse and institutions remain accountable for past wording. We also do not claim that neutralized wording eliminates discriminatory practice. A rewrite should be treated as a signal for substantive institutional review, including review of policy rules, risk indicators, and administrative procedures.

The normative resources used by \model{}, including watchlists, criteria, and protected-group standpoints, are not fixed truths. They should be transparently documented, versioned, periodically audited, and revised with input from affected communities, domain experts, and legal stakeholders. Since LLMs can reproduce social stereotypes, model confidence should not be treated as correctness, and high-risk outputs require manual verification, error analysis, and monitoring for disparate impact.

Finally, the system has dual-use risks. An unrestricted rewriting tool could be misused to sanitize discriminatory policy language without changing the underlying policy. For this reason, we release the code and prompts as research artifacts rather than as a turnkey tool for live administrative deployment, and we do not recommend use in fully automated government decision pipelines.

\section*{Acknowledgments}
This work was supported by the National College Student Innovation and Entrepreneurship Training Program under Project No.~202619145026; the Open Project Program of Marine Ecological Restoration and Smart Ocean Engineering Research Center of Hebei Province under Grant No.~HBMESO2507; the Shijiazhuang--Northeastern University Science and Technology Cooperation Special Project under Grant No.~NEUS2025-01-003; and the Scientific Research Project of Hebei Education Department under Grant No.~QN2024167.

\bibliography{custom}

\appendix

\section*{Appendix Index}
\phantomsection
\label{app:index}

\begin{compactenum}
    \item \hyperref[app:dataset_details]{Dataset Overview}
    \hfill (p.~\pageref{app:dataset_details})

    \item \hyperref[app:extended_benchmarks]{Extended Benchmarks across Model Families}
    \hfill (p.~\pageref{app:extended_benchmarks})
    \begin{compactitem}
        \item \hyperref[app:extended_analysis]{Analysis of Extended Results}
        \hfill (p.~\pageref{app:extended_analysis})
        \item \hyperref[app:rq3_lang]{Cross-Lingual Reasoning} % 修正了这里原本重复的引用标签
        \hfill (p.~\pageref{app:rq3_lang})
    \end{compactitem}

    \item \hyperref[app:loco_texts]{LOCO Prototypes}
    \hfill (p.~\pageref{app:loco_texts})
    \begin{compactitem}
        \item \hyperref[app:loco_prototypes]{Fixed Prototype Texts for Deterministic Mapping}
        \hfill (p.~\pageref{app:loco_prototypes})
    \end{compactitem}

    \item \hyperref[app:end_to_end_traces]{Representative End-to-End Decision Traces}
    \hfill (p.~\pageref{app:end_to_end_traces})

    \item \hyperref[app:rag_details]{Knowledge Base Construction and RAG Strategies}
    \hfill (p.~\pageref{app:rag_details})
    \begin{compactitem}
        \item \hyperref[app:rag_paradigms]{DGDB NEC Retrieval Setting}
        \hfill (p.~\pageref{app:rag_paradigms})
        \item \hyperref[app:kb_composition]{Knowledge Base Composition}
        \hfill (p.~\pageref{app:kb_composition})
        \item \hyperref[app:rag_impl]{Implementation Parameters}
        \hfill (p.~\pageref{app:rag_impl})
    \end{compactitem}

    \item \hyperref[app:hyperparam]{ARG Hyperparameter Sensitivity Grid}
    \hfill (p.~\pageref{app:hyperparam})
    \begin{compactitem}
        \item \hyperref[app:hyperparam_heatmap]{Performance Heatmaps (F1 vs. OGR)}
        \hfill (p.~\pageref{app:hyperparam_heatmap})
    \end{compactitem}

    \item \hyperref[app:prompt_templates]{English and Dutch Prompt Templates}
    \hfill (p.~\pageref{app:prompt_templates})

\end{compactenum}

\vspace{0.8em}

\etocsetnexttocdepth{subsection} 

\section{Dataset Overview}
\label{app:dataset_details}

Tables~\ref{tab:dataset_cards} and~\ref{tab:dataset_cards_external} summarize the datasets used in the experiments.

\begin{table}[h!]
\centering
\footnotesize
\setlength{\tabcolsep}{3pt}
\begin{tabularx}{\columnwidth}{lXX}
\toprule
\rowcolor{cat_zero}
\textbf{Statistics} & \textbf{DGDB} & \textbf{DGDB-Unseen} \\
\midrule
Language & Dutch & Dutch \\
Domain & Gov. docs & Gov. docs \\
Unit & Sentence & Sentence \\
Corpus size & 3,747 & DGDB-derived \\
Evaluation size & Test split & Held-out subset \\
Construction & Original split & Bias-term substitution \\
Evaluation balance & Original split & Label-preserving \\
Targets / labels & 9 bias cats. & Same 9 bias cats. \\
\bottomrule
\end{tabularx}
\caption{Statistical overview of the DGDB-derived datasets.}
\label{tab:dataset_cards}
\end{table}

\begin{table}[h!]
\centering
\scriptsize
\setlength{\tabcolsep}{2pt}
\begin{tabularx}{\columnwidth}{lXXX}
\toprule
\rowcolor{cat_zero}
\textbf{Statistics} & \textbf{DALC} & \textbf{SBIC} & \textbf{KoBBQ} \\
\midrule
Language & Dutch & English & Korean \\
Domain & Twitter & Social & QA \\
Unit & Tweet & Post / tuple & MCQ \\
Corpus size & 8,156 & 44,671 / 147,139 & 76,048 \\
Evaluation size & 1,000 & 1,000 & 1,000 \\
Evaluation balance & 500 / 500 & 500 / 500 & 500 / 500 \\
Targets / labels & Abusive binary & Bias tuple & Bias contexts \\
\bottomrule
\end{tabularx}
\caption{Statistical overview of the external evaluation datasets.}
\label{tab:dataset_cards_external}
\end{table}

\section{Extended Benchmarks across Model Families}
\label{app:extended_benchmarks}

This appendix tests whether \model{}'s gains depend on a single frontier backbone. We evaluate 18 LLMs, grouped into \textit{Frontier SOTA}, \textit{Strong Generalist}, and \textit{Mid/Efficiency} tiers. Table~\ref{tab:extended_results} reports the corresponding zero-shot and \model{} results on DGDB.

\begin{table*}[ht!]
    \centering
    \small % 使用字体大小控制而非过度缩放，保持文字清晰
    \resizebox{0.98\textwidth}{!}{%
    \begin{tabular}{l|l|ccc|ccc|c}
    \toprule
    \multirow{2}{*}{\textbf{Model Family}} & \multirow{2}{*}{\textbf{Model Name}} & \multicolumn{3}{c|}{\textbf{Standard Zero-shot}} & \multicolumn{3}{c|}{\textbf{Ours: \model{}}} & \textbf{Governance} \\
     & & \textbf{Prec} & \textbf{Rec} & \textbf{F1} & \textbf{Prec} & \textbf{Rec} & \textbf{F1} & \textbf{BGS} \\ \midrule
    \multirow{5}{*}{\textit{Frontier SOTA}}
     & \textbf{Gemini 2.5 Pro} & 0.524\,\scriptsize{$\pm$.024} & 0.907\,\scriptsize{$\pm$.018} & 0.664\,\scriptsize{$\pm$.016} & \textbf{0.865}\,\scriptsize{$\pm$.015} & 0.895\,\scriptsize{$\pm$.013} & \textbf{0.880}\,\scriptsize{$\pm$.012} & \textbf{0.808}\,\scriptsize{$\pm$.013} \\
     & DeepSeek-R1 & 0.549\,\scriptsize{$\pm$.023} & 0.887\,\scriptsize{$\pm$.019} & 0.678\,\scriptsize{$\pm$.015} & \textbf{0.872}\,\scriptsize{$\pm$.016} & 0.884\,\scriptsize{$\pm$.014} & 0.878\,\scriptsize{$\pm$.013} & 0.806\,\scriptsize{$\pm$.015} \\
     & Claude 3.5 Sonnet & \textbf{0.562}\,\scriptsize{$\pm$.022} & 0.872\,\scriptsize{$\pm$.020} & \textbf{0.683}\,\scriptsize{$\pm$.016} & 0.860\,\scriptsize{$\pm$.017} & 0.882\,\scriptsize{$\pm$.015} & 0.871\,\scriptsize{$\pm$.013} & 0.787\,\scriptsize{$\pm$.016} \\
     & Grok-3 & 0.484\,\scriptsize{$\pm$.031} & \textbf{0.920}\,\scriptsize{$\pm$.017} & 0.634\,\scriptsize{$\pm$.019} & 0.836\,\scriptsize{$\pm$.019} & \textbf{0.910}\,\scriptsize{$\pm$.013} & 0.871\,\scriptsize{$\pm$.014} & 0.779\,\scriptsize{$\pm$.018} \\
     & GPT-4o & 0.514\,\scriptsize{$\pm$.025} & 0.882\,\scriptsize{$\pm$.021} & 0.649\,\scriptsize{$\pm$.017} & 0.844\,\scriptsize{$\pm$.017} & 0.863\,\scriptsize{$\pm$.015} & 0.853\,\scriptsize{$\pm$.013} & 0.757\,\scriptsize{$\pm$.014} \\ \midrule
    \multirow{6}{*}{\textit{Strong Generalist}}
     & Claude 3 Opus & 0.531\,\scriptsize{$\pm$.026} & 0.872\,\scriptsize{$\pm$.021} & 0.660\,\scriptsize{$\pm$.018} & 0.852\,\scriptsize{$\pm$.018} & 0.867\,\scriptsize{$\pm$.016} & 0.859\,\scriptsize{$\pm$.014} & 0.749\,\scriptsize{$\pm$.018} \\
     & GPT-4 Turbo & 0.494\,\scriptsize{$\pm$.031} & 0.862\,\scriptsize{$\pm$.020} & 0.628\,\scriptsize{$\pm$.021} & 0.836\,\scriptsize{$\pm$.019} & 0.852\,\scriptsize{$\pm$.018} & 0.844\,\scriptsize{$\pm$.015} & 0.717\,\scriptsize{$\pm$.017} \\
     & DeepSeek-V3 & 0.482\,\scriptsize{$\pm$.033} & 0.846\,\scriptsize{$\pm$.023} & 0.614\,\scriptsize{$\pm$.020} & 0.811\,\scriptsize{$\pm$.022} & 0.832\,\scriptsize{$\pm$.019} & 0.821\,\scriptsize{$\pm$.017} & 0.682\,\scriptsize{$\pm$.021} \\
     & Qwen-2.5-72B & 0.472\,\scriptsize{$\pm$.036} & 0.852\,\scriptsize{$\pm$.024} & 0.608\,\scriptsize{$\pm$.023} & 0.819\,\scriptsize{$\pm$.021} & 0.839\,\scriptsize{$\pm$.019} & 0.829\,\scriptsize{$\pm$.016} & 0.696\,\scriptsize{$\pm$.019} \\
     & Mistral Large 2 & 0.491\,\scriptsize{$\pm$.034} & 0.828\,\scriptsize{$\pm$.026} & 0.616\,\scriptsize{$\pm$.022} & 0.806\,\scriptsize{$\pm$.022} & 0.826\,\scriptsize{$\pm$.019} & 0.816\,\scriptsize{$\pm$.018} & 0.654\,\scriptsize{$\pm$.022} \\
     & Llama-3-70B & 0.463\,\scriptsize{$\pm$.037} & 0.837\,\scriptsize{$\pm$.024} & 0.596\,\scriptsize{$\pm$.023} & 0.800\,\scriptsize{$\pm$.023} & 0.820\,\scriptsize{$\pm$.020} & 0.810\,\scriptsize{$\pm$.017} & 0.668\,\scriptsize{$\pm$.023} \\
     \noalign{\smallskip}
     \arrayrulecolor{red}\cdashline{1-9}[3pt/2pt] \arrayrulecolor{black}
     \noalign{\smallskip}
    \multicolumn{9}{c}{\textit{\textcolor{red}{\small --- Performance Threshold: Fine-tuned BERTje (F1 = 0.812) ---}}} \\ \midrule
    \multirow{7}{*}{\textit{Mid/Efficiency}}
     & Gemini 1.5 Pro & 0.478\,\scriptsize{$\pm$.038} & 0.821\,\scriptsize{$\pm$.027} & 0.604\,\scriptsize{$\pm$.024} & 0.781\,\scriptsize{$\pm$.024} & 0.798\,\scriptsize{$\pm$.021} & 0.789\,\scriptsize{$\pm$.018} & 0.612\,\scriptsize{$\pm$.022} \\
     & Qwen-2.5-32B & 0.422\,\scriptsize{$\pm$.041} & 0.798\,\scriptsize{$\pm$.028} & 0.552\,\scriptsize{$\pm$.026} & 0.742\,\scriptsize{$\pm$.026} & 0.762\,\scriptsize{$\pm$.023} & 0.752\,\scriptsize{$\pm$.019} & 0.575\,\scriptsize{$\pm$.024} \\
     & Claude 3 Haiku & 0.412\,\scriptsize{$\pm$.043} & 0.778\,\scriptsize{$\pm$.031} & 0.539\,\scriptsize{$\pm$.029} & 0.718\,\scriptsize{$\pm$.027} & 0.738\,\scriptsize{$\pm$.023} & 0.728\,\scriptsize{$\pm$.021} & 0.539\,\scriptsize{$\pm$.027} \\
     & Gemini 1.5 Flash & 0.401\,\scriptsize{$\pm$.042} & 0.789\,\scriptsize{$\pm$.030} & 0.532\,\scriptsize{$\pm$.027} & 0.703\,\scriptsize{$\pm$.027} & 0.723\,\scriptsize{$\pm$.025} & 0.713\,\scriptsize{$\pm$.021} & 0.521\,\scriptsize{$\pm$.027} \\
     & Mixtral 8x7B & 0.382\,\scriptsize{$\pm$.045} & 0.762\,\scriptsize{$\pm$.033} & 0.509\,\scriptsize{$\pm$.031} & 0.671\,\scriptsize{$\pm$.029} & 0.694\,\scriptsize{$\pm$.026} & 0.682\,\scriptsize{$\pm$.023} & 0.486\,\scriptsize{$\pm$.029} \\
     & GPT-3.5 Turbo & 0.352\,\scriptsize{$\pm$.048} & 0.748\,\scriptsize{$\pm$.036} & 0.479\,\scriptsize{$\pm$.033} & 0.633\,\scriptsize{$\pm$.032} & 0.654\,\scriptsize{$\pm$.027} & 0.643\,\scriptsize{$\pm$.024} & 0.448\,\scriptsize{$\pm$.031} \\
     & Llama-3-8B & 0.321\,\scriptsize{$\pm$.051} & 0.712\,\scriptsize{$\pm$.037} & 0.442\,\scriptsize{$\pm$.035} & 0.591\,\scriptsize{$\pm$.033} & 0.611\,\scriptsize{$\pm$.029} & 0.601\,\scriptsize{$\pm$.026} & 0.397\,\scriptsize{$\pm$.033} \\ \bottomrule
    \end{tabular}%
    }
    \caption{\textbf{Extended Leaderboard on DGDB (In-Domain).} 
    Models are ranked by their \model{} F1 score. 
    The red dashed line marks the performance threshold of the fine-tuned BERTje baseline ($F1=0.812$). 
    \textbf{Observation:} While \model{} improves performance across model families (avg. $\Delta$ F1 $+0.20$, larger for frontier backbones than for smaller ones), a minimum reasoning capability threshold (approx. GPT-4 Turbo level) is required to surpass supervised baselines.}
\label{tab:extended_results}
\end{table*}

\subsection{Analysis of Extended Results}
\label{app:extended_analysis}
The extended results show broad gains across backbones, but the final ceiling still depends on reasoning capacity. Models at or above the GPT-4 Turbo / Llama-3-70B tier reliably exceed the fine-tuned BERTje baseline, including open-weight options such as \textit{Qwen-2.5-72B} and \textit{DeepSeek-V3}.

\subsection{Cross-Lingual Reasoning}
\label{app:rq3_lang}

\begin{figure}[h]
    \centering
    \includegraphics[width=0.8\columnwidth]{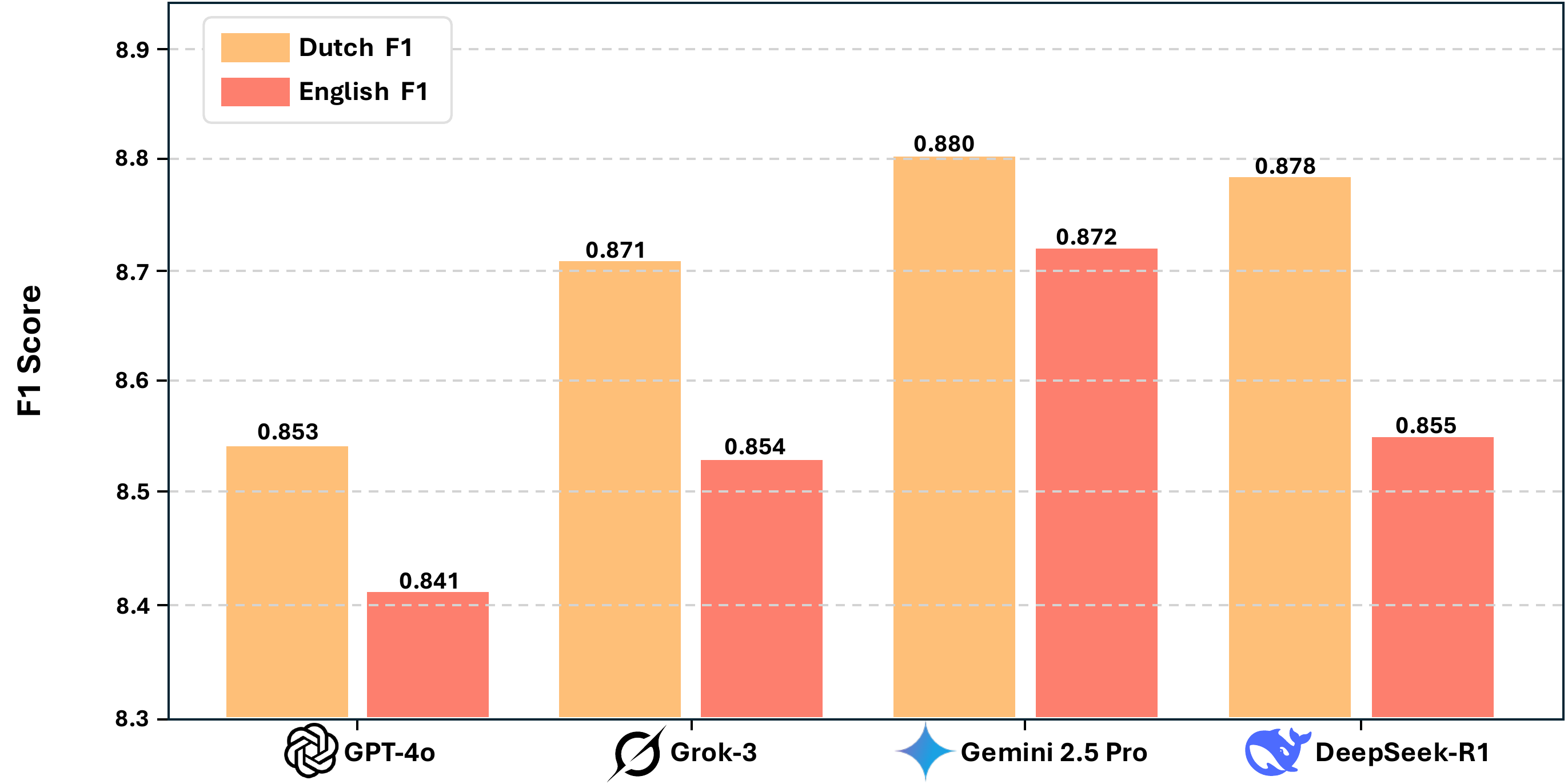}
    \caption{Comparison of F1 scores between Native Dutch and Translated English prompts. Native prompting consistently outperforms English across all backbones, avoiding the semantic loss associated with translating specific administrative terminology ($p < 0.05$, paired Student's t-test over 3 independent runs).}
\label{fig:lang_comparison}
\end{figure}

\noindent\textbf{Native vs. English Reasoning.}
Figure~\ref{fig:lang_comparison} compares Dutch prompting with English prompting after DeepL translation. All scores are averaged over 3 independent runs.

Dutch prompting gives a small but consistent F1 advantage across all four backbones ($+0.8\%$ to $+2.3\%$), mainly because administrative terms such as \textit{"afstand tot de arbeidsmarkt"} and \textit{"leefvorm"} lose part of their meaning after translation.

\section{LOCO Prototypes}
\label{app:loco_texts}

To support reproducibility of the deterministic open-set evaluation, we provide the full fixed prototype texts $e(c)$ used for cosine-similarity mapping. LOCO uses \texttt{clips/e5-base-trm-nl} for prototype ranking, whereas NEC uses the separate retrieval embedding model described in Appendix~\ref{app:rag_impl}.

\subsection{Fixed Prototype Texts for Deterministic Mapping}
\label{app:loco_prototypes}
The following prototype vectors were fixed before the experiment based on dataset statistics and definitions. The Scout's generated \texttt{group} and \texttt{rationale} were mapped against these to calculate Correct@1 and Correct@3.

\begin{itemize}[leftmargin=*, noitemsep]
    \item \textbf{B1 (Disability):} \textit{"beperkingen, handicap, rolstoelgebonden, dwerg, doventolk, medische toestand, fysieke of mentale beperking."}
    \item \textbf{B2 (Migration):} \textit{"migratie, allochtoon, vluchteling, nieuwkomer, asielzoeker, integratie, westers, niet-westers."}
    \item \textbf{B3 (Colonialism):} \textit{"kolonialisme, slavernijverleden, inheems, etnisch, blank, zwart, ras, exotisch, primitief."}
    \item \textbf{B4 (Religion):} \textit{"geloof, religie, christen, moslim, islam, joods, hoofddoek, extremisme."}
    \item \textbf{B5 (Gender):} \textit{"gender, geslacht, man, vrouw, meisje, jongen, seksisme, traditionele rolpatronen."}
    \item \textbf{B6 (LGBTQ+):} \textit{"seksualiteit, geaardheid, homo, hetero, queer, transgender, non-binair, lhbtiq+."}
    \item \textbf{B7 (Culture):} \textit{"cultuur, traditie, bi-cultureel, tussenpositie, culturele achtergrond, minderheden."}
    \item \textbf{B8 (Education/Class):} \textit{"onderwijs, klasse, achterstandsleerling, mbo, laagopgeleid, probleemwijk, sociale status."}
    \item \textbf{B9 (Welfare):} \textit{"algemeen, armoede, machtsongelijkheid, fobie, sociaal-economische kwetsbaarheid."}
\end{itemize}

\section{Representative End-to-End Decision Traces}
\label{app:end_to_end_traces}

We provide three representative traces that illustrate the main decision paths: severe intersectional bias, false inference from language proficiency, and a neutral procedural case. These traces explain the decision mechanism; aggregate detection and ablation results are reported in Tables~\ref{tab:main_results_heatmap} and~\ref{tab:ablation}. The full prompt definitions used to produce these traces are available in \texttt{agent\_bias/common/prompts.py}.

\begin{tracecase}{Case T1: Intersectional welfare-fraud profiling}
\textbf{Input.} \textit{"Om uitkeringsfraude tegen te gaan, moet de gemeente strengere huisbezoeken uitvoeren bij allochtone alleenstaande moeders in de Schilderswijk, aangezien deze groep historisch gezien een gebrek aan administratief inzicht heeft."}
\tcblower
\textbf{Trace.} The evidence layer retrieves anti-discrimination and profiling guidance. The Scout identifies migration, gender, and neighborhood/class proxies. Jurors for migration, gender, class, and welfare flag the sentence because it links protected and socio-economic attributes to fraud risk and administrative incapacity.

\textbf{Decision.} The Judge returns \textsc{Biased}; the rewrite preserves the fraud-control goal but replaces group targeting with objective risk indicators.
\end{tracecase}

\begin{tracecase}{Case T2: Language proficiency as intent inference}
\textbf{Input.} \textit{"De aanvrager toonde tijdens het gesprek een gebrekkige beheersing van de Nederlandse taal; dit duidt op een gebrek aan integratie en een verhoogde kans op opzettelijke misleiding."}
\tcblower
\textbf{Trace.} The Scout identifies non-native speakers as the affected group. Jurors flag the unsupported causal chain from language ability to integration failure and intentional deception.

\textbf{Decision.} The Judge returns \textsc{Biased}; the rewrite reframes the observation as a communication-support need while preserving the administrative fact.
\end{tracecase}

\begin{tracecase}{Case T3: Objective income threshold}
\textbf{Input.} \textit{"Om in aanmerking te komen voor de huurtoeslag, mag het gezamenlijke toetsingsinkomen niet hoger zijn dan €32.500 per jaar."}
\tcblower
\textbf{Trace.} The evidence layer identifies a statutory eligibility threshold. The Scout finds no protected or stigmatized group target beyond the general applicant population. Jurors treat the sentence as a neutral criterion.

\textbf{Decision.} The Judge returns \textsc{Non-Biased}, and restoration is skipped.
\end{tracecase}

\section{Knowledge Base Construction and RAG Strategies}
\label{app:rag_details}

This appendix summarizes the NEC knowledge base $\mathcal{K}$ and retrieval settings used by \model{}.

\subsection{DGDB NEC Retrieval Setting}
\label{app:rag_paradigms}

For DGDB, NEC retrieval uses the source inventory released with the code. It combines legal and policy foundations, accessibility and administrative communication guidance, reports on online discrimination and institutional racism, inclusive-language resources, and DGDB category reasoning notes.

\subsection{Knowledge Base Composition}
\label{app:kb_composition}
The curated repository $\mathcal{K}$ is structured into three functional layers:
\begin{itemize}
    \item \textbf{Legal and Policy Core (Grondslag):} \textit{Grondwet} Chapter 1, AWGB core articles, OCW antidiscrimination agenda material, the Dutch national programme against discrimination and racism, and ECRI Netherlands extracts.
    \item \textbf{Administrative and Accessibility Guides (Overheidsrichtlijnen):} accessible voting plans, election communication guidance, and official inclusive and accessible writing guides.
    \item \textbf{Sociocultural and Inclusive-Language Resources (Inclusiviteit):} online-discrimination and harmful-behavior reports, institutional-racism summaries, emancipation monitoring, DEBIAS vocabulary entries, inclusive communication glossaries, cultural-sector language guidance, and DGDB reasoning entries.
\end{itemize}
Table~\ref{tab:rag_strategy_comparison} summarizes how this inventory supports the DGDB retrieval setting.

\begin{table}[h!]
\centering
\small
\resizebox{\columnwidth}{!}{%
\begin{tabular}{l|p{5.5cm}}
\toprule
\textbf{Dimension} & \textbf{DGDB NEC retrieval} \\ \midrule
\textbf{Primary Objective} & Detecting implicit bias in official policy texts. \\
\textbf{Core Domain} & Legal and political administrative text (\textit{high-stakes}). \\ \midrule
\textbf{Inventory} & Dutch constitutional and equal-treatment law; antidiscrimination policy; accessibility and election communication guidance; online-discrimination, institutional-racism, emancipation, and inclusive-language resources; DGDB reasoning entries. \\ \midrule
\textbf{Selection Rationale} & Grounds the review in legal norms while covering administrative, accessibility, and inclusive-language contexts that commonly shape implicit bias. \\ \bottomrule
\end{tabular}%
}
\caption{\textbf{NEC retrieval construction for DGDB.} \model{} prioritizes normative sources and official writing guidance for administrative bias review.}
\label{tab:rag_strategy_comparison}
\end{table}

\subsection{Implementation Parameters}
\label{app:rag_impl}
NEC utilizes \texttt{text-embedding-3-small} for RAG dense retrieval. This is separate from the \texttt{clips/e5-base-trm-nl} encoder used for semantic fidelity and LOCO prototype ranking. The released implementation first retrieves keyword matches, then dense semantic matches, and finally expands selected hits to adjacent chunks from the same source. Long policy documents use per-source character windows, while DGDB reasoning entries are split by markdown headings. At inference, the default setting selects $k=5$ seed fragments before local context expansion, ensuring the Summary Agent synthesizes a contextually grounded, neutral evidence base.

The three retrieval stages are complementary: keyword matching preserves precise legal and administrative terminology, dense retrieval recovers paraphrases, and local expansion restores nearby definitions or caveats. Source identifiers and chunk positions are retained in the decision trace, allowing the retrieved evidence to be inspected independently of the verdict. The Summary Agent condenses this material but cannot classify the sentence, keeping retrieval errors separate from later reasoning errors.

\section{ARG Hyperparameter Sensitivity Grid}
\label{app:hyperparam}

We tune ARG on the validation set with Gemini 2.5 Pro. Five parameters govern escalation: $\tau_{\mathrm{high}}{=}0.6$ immediately flags \texttt{HARD} cases, $\tau_{\mathrm{high\_risk}}{=}0.7$ applies a stricter watchlist threshold, $\tau_{\mathrm{mid}}{=}0.7$ triggers \texttt{SOFT} jury review, $N_{\mathrm{soft\_bias}}{=}2$ confirms bias in \texttt{SOFT} mode, and $N_{\mathrm{soft\_poss}}{=}1$ marks possible bias for manual review. The heatmap varies the two most influential parameters: $\tau_{\mathrm{high}}$ and $N_{\mathrm{soft\_bias}}$.

\subsection{Performance Heatmaps}
\label{app:hyperparam_heatmap}

\begin{figure}[ht]
  \centering
  \includegraphics[width=0.92\linewidth]{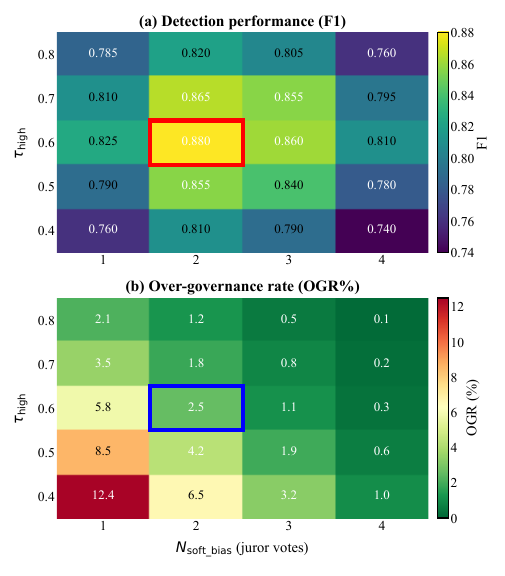}
  \vspace{-0.45em}
  \caption{\textbf{ARG Hyperparameter Grid Search.} (a) F1 Score and (b) Over-Governance Rate (OGR) across varying $\tau_{\mathrm{high}}$ and juror-vote thresholds. The selected configuration ($\tau_{\mathrm{high}}=0.6, N_{\mathrm{soft\_bias}}=2$) is highlighted.}
\label{fig:arg_heatmap}
\end{figure}

Figure \ref{fig:arg_heatmap} illustrates the interaction between these ARG parameters.

\paragraph{Selected Point.}
Low thresholds bypass the jury too often, while high thresholds increase calls without improving F1. The selected setting ($\tau_{\mathrm{high}}=0.6, N_{\mathrm{soft\_bias}}=2$) reaches F1 $=0.880$ with OGR $=2.5\%$.

\section{English and Dutch Prompt Templates}
\label{app:prompt_templates}

This appendix gives the English and Dutch prompt templates for the main \model{} agents. The templates are formatted for readability; placeholders such as \pfield{sentence}, \pfield{evidence}, and \pfield{jury\_outputs} denote runtime fields. Prompt boundaries mirror the pipeline stages and restrict each agent to a specific role and output contract. Evidence summarization, standpoint discovery, adjudication, restoration, and verification therefore remain separately inspectable. Malformed outputs are exposed to the routing logic rather than interpreted heuristically.

\subsection{English Prompt Templates}

\begin{promptbox}{P1. NEC Summary Agent}
\prompttag{System role.}
You are a context summarizer for governance text screening. Select and summarize only the context fragments that are directly relevant to the input sentence. Be objective and factual: no verdict, no diagnosis, no classification, and no interpretation of intent. Use only information that follows literally or unambiguously from the supplied context.

\tcblower
\prompttag{User task.}
Given \pfield{query} and retrieved \pfield{context}, produce a short neutral evidence summary before any bias decision is made.

\prompttag{Output contract.}
Return \textbf{3--6 bullets}. Each bullet must contain one paraphrased context fragment and one neutral relevance label: \textit{definition}, \textit{criterion}, \textit{guideline}, \textit{example}, \textit{caveat}, \textit{scope}, or \textit{terminology}. If no relevant evidence exists, return exactly one bullet stating that no sufficiently relevant context was found. Do not mention source names, links, file names, or citations.
\end{promptbox}

\begin{promptbox}{P2. Scout Agent for Open-Set Standpoint Screening}
\prompttag{System role.}
Identify which social groups may be referenced or affected by the sentence. Focus on groups that are explicitly mentioned or strongly implied, especially groups outside the fixed nine-category panel. Do not judge whether the sentence is biased; only detect and describe affected groups.

\prompttag{Known panel.}
B1 disability/functional limitations; B2 migration/refugees; B3 colonial history/slavery narratives; B4 religion, especially Islam; B5 gender and gender roles; B6 LGBTQ+; B7 cultural or ethnic groups; B8 education/class; B9 welfare and social vulnerability.

\prompttag{Policy-language cues.}
Treat context-sensitive expressions as possible group markers when they function that way in administrative text. Examples include \textit{parents} in education/youth-policy contexts, \textit{background} or \textit{minorities} for cultural/ethnic groups, religious identifiers such as \textit{Islam} or \textit{headscarf}, and migration terms such as \textit{newcomer} or \textit{non-Western}.

\tcblower
\prompttag{User task.}
For \pfield{sentence}, return exactly one JSON object:
\begin{quote}\ttfamily\footnotesize
\{ "detected\_groups": [ \{ "group\_label": "...", "is\_known\_category": true/false, "mapped\_jury": "B1|...|B9|none", "evidence": ["..."], "x\_prompt\_hint": "..." \} ] \}
\end{quote}
\promptwarn{Constraints.}
Use short generic group labels; add multiple objects when multiple groups are implicated; write \texttt{x\_prompt\_hint} only for a newly surfaced group; return an empty list when no relevant group is found.
\end{promptbox}

\begin{promptbox}{P3. Juror Agents}
\prompttag{Common user prompt.}
Each juror receives \pfield{sentence} and the NEC evidence summary \pfield{evidence}. The juror returns exactly one JSON object with \texttt{agent\_id}, \texttt{group\_labels}, \texttt{label}, \texttt{confidence}, \texttt{bias\_signals}, \texttt{analysis}, and \texttt{risk\_score}.

\prompttag{Label contract.}
\textbf{Biased} means negative framing, stereotyping, exclusion, or dehumanization directed at the juror's target group. \textbf{PossiblyBiased} means a concrete textual cue exists but evidence is not decisive. \textbf{Non-biased} means neutral, factual, or positive toward the target group. \textbf{Uncertain} means insufficient or ambiguous evidence.

\promptwarn{Grounding rules.}
Do not label a sentence biased solely because a watchlist term appears. The juror must cite a short signal from the sentence or evidence summary. If no signal can be cited, use \textbf{Uncertain}. Calibrate \texttt{risk\_score}: high for prohibited language or direct threat/criminality links, medium for clear negative framing, and low for weak or doubtful signals.

\tcblower
\prompttag{Dynamic 10th juror template.}
You represent \pfield{group\_label}. Flag possible open-set bias for this group even when other jurors consider the risk low. Look for subtle exclusion, stereotyping, marginalization, presupposition, neutral-sounding policy frames, implicit causality, group generalizations, and loaded labels. Use only the sentence and evidence summary; if evidence is weak, choose \textbf{Uncertain} and state what is missing. Runtime hint: \pfield{rationale}.
\end{promptbox}

\begin{promptbox}{P4. Fixed Juror Standpoints}
The fixed jurors share P3's output contract but differ in standpoint:
\begin{compactitem}
    \item \textbf{B1 Disability:} watch for disability framed as burden, cost, risk, dependency, incapacity, or exclusion from participation.
    \item \textbf{B2 Migration:} watch for massing metaphors, threat framing, securitization, blame, and unsupported links to disorder, abuse, or incompatibility.
    \item \textbf{B3 Colonialism:} watch for civilizing frames, gratitude or debt frames, agency erasure, euphemisms, and softened accounts of coercion or exploitation.
    \item \textbf{B4 Religion:} watch for collective blame, essentialism, links between religion and threat, and problematization of religious practice without nuance.
    \item \textbf{B5 Gender:} watch for stereotyped roles, male-default norms, objectification, vulnerability framing, and positive stereotypes that preserve hierarchy.
    \item \textbf{B6 LGBTQ+:} watch for hetero/cisnormativity, pathologization, moral-panic framing, and neutral-sounding exclusion from rights or services.
    \item \textbf{B7 Culture/Ethnicity:} watch for essentialist culture claims, us/them framing, exoticization, hierarchy, and ethnicity linked to threat or deviance.
    \item \textbf{B8 Education/Class:} watch for condescension, status hierarchy, deficit assumptions, and access rules that treat higher-status participation as default.
    \item \textbf{B9 Welfare:} watch for burden framing, victim blaming, criminalization, dehumanization, and treating poverty, unemployment, or homelessness as uniform stereotypes.
\end{compactitem}
\end{promptbox}

\begin{promptbox}{P5. Judge Agent}
\prompttag{System role.}
Decide the final binary label for linguistic bias in a Dutch policy-style sentence: biased (1) or non-biased (0). Apply the definition strictly: a sentence is biased if and only if \textbf{Condition A} holds, or \textbf{Condition B1 and Condition B2} both hold.

\prompttag{Condition A: prohibited terms.}
The sentence contains a universally inappropriate term, unless it is a direct quotation or is explicitly used to explain or criticize harmful language.

\prompttag{Condition B1: contextual bias.}
The sentence exhibits stereotyping, exclusion, power imbalance, or prejudice, either explicit or implicit.

\prompttag{Condition B2: directed at a group.}
The bias is directed at a specific social group that is explicitly mentioned or unambiguously implied.

\tcblower
\prompttag{User task.}
Use only \pfield{sentence}, \pfield{evidence}, and raw \pfield{jury\_outputs}. Prioritize quoted language signals and consistent multi-agent signals over vague impressions; this is not simple majority voting. Use confidence and risk scores only as weighting hints.

\prompttag{Output contract.}
Return exactly one JSON object:
\begin{quote}\ttfamily\footnotesize
\{ "label": 0/1, "confidence": 0--1, "rationale": "max 3 sentences; cite decisive signals/agents" \}
\end{quote}
When jurors conflict, prefer quoted prohibited terms, dehumanizing or threat/burden framing, and higher-confidence analyses only when they are grounded in concrete signals.
\end{promptbox}

\begin{promptbox}{P6. Restoration Agent}
\prompttag{System role.}
Rewrite Dutch policy-style sentences to remove linguistic bias while preserving original meaning, facts, administrative register, and policy intent. Assume the sentence has already been flagged as biased upstream; the task is mitigation, not reclassification.

\prompttag{Edit policy.}
Make the smallest possible edits that neutralize the bias: replace prohibited or problematic terms, remove stigmatizing metaphors, remove stereotyping or dehumanizing framing, and use neutral or person-first wording where appropriate. Keep the output to one formal Dutch sentence. Do not add new facts, actors, numbers, causal claims, or assumptions.

\tcblower
\prompttag{User task.}
Given \pfield{sentence}, \pfield{evidence}, \pfield{aggregate}, \pfield{orig\_label}, and \pfield{sample\_id}, return exactly one JSON object:
\begin{quote}\ttfamily\footnotesize
\{ "rewritten\_text": "...", "new\_label": 0, "original\_label": \pfield{orig\_label}, "id": "\pfield{sample\_id}", "notes": "1--2 key edits" \}
\end{quote}
\promptwarn{Fidelity rule.}
Preserve all concrete facts from the original sentence unless the factual wording itself is the biased framing. The rewrite must not become an explanation, bullet list, or multi-sentence paraphrase.
\end{promptbox}

\subsection{Dutch Prompt Templates}

\begin{dutchpromptbox}{D1. NEC Summary Agent}
\prompttag{Systeemrol.}
Je bent een Nederlandstalige context-samenvatter voor screening van bestuursteksten. Selecteer en vat alleen contextfragmenten samen die direct relevant zijn voor de gegeven zin. Werk strikt objectief en feitelijk: geen oordeel, geen diagnose, geen classificatie en geen interpretatie van intentie. Gebruik alleen informatie die letterlijk of ondubbelzinnig uit de aangeleverde context volgt.

\tcblower
\prompttag{Gebruikerstaak.}
Gegeven \pfield{query} en de opgehaalde \pfield{context}, maak een korte neutrale evidence-samenvatting voordat een biasbeslissing wordt genomen.

\prompttag{Outputcontract.}
Geef \textbf{3--6 bullets}. Elke bullet bevat een geparafraseerd contextfragment en een neutraal relevantielabel: \textit{definitie}, \textit{criterium}, \textit{richtlijn}, \textit{voorbeeld}, \textit{aandachtspunt}, \textit{nuance/afbakening} of \textit{terminologie}. Als er geen relevante evidence is, geef exact een bullet die meldt dat geen voldoende relevante context is gevonden. Noem geen brondata, links, bestandsnamen of citaties.
\end{dutchpromptbox}

\begin{dutchpromptbox}{D2. Scout Agent voor Open-Set Standpoint Screening}
\prompttag{Systeemrol.}
Identificeer welke sociale groepen mogelijk worden genoemd of geraakt door de zin. Focus op groepen die expliciet genoemd worden of sterk geïmpliceerd zijn, vooral groepen buiten het vaste panel van negen categorieën. Geef geen oordeel over bias; detecteer en beschrijf alleen mogelijke doelgroepen.

\prompttag{Vast panel.}
B1 beperking/functionele beperking; B2 migratie/vluchtelingen; B3 koloniale geschiedenis/slavernijnarratieven; B4 religie, vooral islam; B5 gender en genderrollen; B6 LGBTQ+; B7 culturele of etnische groepen; B8 onderwijs/klasse; B9 welzijn en sociale kwetsbaarheid.

\prompttag{Cues in beleidstaal.}
Behandel contextgevoelige termen als mogelijke groepsmarkeringen wanneer zij zo functioneren in bestuurstekst. Voorbeelden zijn \textit{ouders} in onderwijs- of jeugdbeleid, \textit{achtergrond} of \textit{minderheden} voor culturele/etnische groepen, religieuze markeerders zoals \textit{islam} of \textit{hoofddoek}, en migratietermen zoals \textit{nieuwkomer} of \textit{niet-westers}.

\tcblower
\prompttag{Gebruikerstaak.}
Voor \pfield{sentence}, retourneer exact een JSON-object:
\begin{quote}\ttfamily\footnotesize
\{ "detected\_groups": [ \{ "group\_label": "...", "is\_known\_category": true/false, "mapped\_jury": "B1|...|B9|none", "evidence": ["..."], "x\_prompt\_hint": "..." \} ] \}
\end{quote}
\promptwarn{Beperkingen.}
Gebruik korte generieke groepslabels; voeg meerdere objecten toe wanneer meerdere groepen geraakt worden; vul \texttt{x\_prompt\_hint} alleen voor een nieuw opgekomen groep; retourneer een lege lijst wanneer geen relevante groep wordt gevonden.
\end{dutchpromptbox}

\begin{dutchpromptbox}{D3. Juror Agents}
\prompttag{Gemeenschappelijke gebruikerstaak.}
Elke juror ontvangt \pfield{sentence} en de NEC evidence-samenvatting \pfield{evidence}. De juror retourneert exact een JSON-object met \texttt{agent\_id}, \texttt{group\_labels}, \texttt{label}, \texttt{confidence}, \texttt{bias\_signals}, \texttt{analysis} en \texttt{risk\_score}.

\prompttag{Labelcontract.}
\textbf{Biased} betekent negatieve framing, stereotypering, uitsluiting of ontmenselijking gericht op de doelgroep van de juror. \textbf{PossiblyBiased} betekent dat er een concreet taalsignaal is, maar dat het bewijs nog niet doorslaggevend is. \textbf{Non-biased} betekent neutraal, feitelijk of positief ten opzichte van de doelgroep. \textbf{Uncertain} betekent onvoldoende of ambigue informatie.

\promptwarn{Groundingregels.}
Label een zin niet als biased alleen omdat een watchlistterm voorkomt. De juror moet een kort signaal uit de zin of evidence-samenvatting kunnen citeren. Als dat niet kan, kies \textbf{Uncertain}. Kalibreer \texttt{risk\_score}: high voor verboden taal of directe dreiging/criminaliteitskoppeling, medium voor duidelijke negatieve framing, en low voor zwakke of twijfelachtige signalen.

\tcblower
\prompttag{Dynamische 10e juror-template.}
Jij vertegenwoordigt \pfield{group\_label}. Signaleer mogelijke open-set bias voor deze groep, ook wanneer andere jurors het risico laag inschatten. Let op subtiele uitsluiting, stereotypering, marginalisering, vooronderstellingen, neutraal klinkende beleidsframes, impliciete causaliteit, groepsgeneralisaties en geladen labels. Gebruik alleen de zin en evidence-samenvatting; kies \textbf{Uncertain} wanneer evidence zwak is en benoem wat ontbreekt. Runtime-hint: \pfield{rationale}.
\end{dutchpromptbox}

\begin{dutchpromptbox}{D4. Vaste Juror-standpunten}
De vaste jurors delen het outputcontract van D3, maar verschillen in standpoint:
\begin{compactitem}
    \item \textbf{B1 Beperking:} let op beperking als last, kostenpost, risico, afhankelijkheid, onbekwaamheid of uitsluiting van participatie.
    \item \textbf{B2 Migratie:} let op massametaforen, dreigingsframes, securitisering, schuldtoewijzing en ongefundeerde koppelingen aan wanorde, misbruik of onverenigbaarheid.
    \item \textbf{B3 Kolonialisme:} let op beschavingsframes, dankbaarheids- of schuldframes, agency-uitwissing, eufemismen en verzachte beschrijvingen van dwang of uitbuiting.
    \item \textbf{B4 Religie:} let op collectieve schuld, essentialisme, koppelingen tussen religie en dreiging, en problematisering van religieuze praktijk zonder nuance.
    \item \textbf{B5 Gender:} let op stereotiepe rollen, mannelijke standaardnormen, objectificatie, kwetsbaarheidsframing en positieve stereotypen die hiërarchie bewaren.
    \item \textbf{B6 LGBTQ+:} let op hetero/cisnormativiteit, pathologisering, morele-paniekframing en neutraal klinkende uitsluiting van rechten of diensten.
    \item \textbf{B7 Cultuur/Etniciteit:} let op essentialistische cultuurclaims, wij/zij-framing, exotisering, hiërarchie en etniciteit gekoppeld aan dreiging of deviant gedrag.
    \item \textbf{B8 Onderwijs/Klasse:} let op neerbuigendheid, statushiërarchie, deficit-aannames en toegangseisen die hogere-statusdeelname als norm behandelen.
    \item \textbf{B9 Welzijn:} let op lastframing, victim blaming, criminalisering, ontmenselijking en het behandelen van armoede, werkloosheid of dakloosheid als uniform stereotype.
\end{compactitem}
\end{dutchpromptbox}

\begin{dutchpromptbox}{D5. Judge Agent}
\prompttag{Systeemrol.}
Beslis het finale binaire label voor taalkundige bias in een Nederlandse beleid-achtige zin: biased (1) of non-biased (0). Pas de definitie strikt toe: een zin is biased als en alleen als \textbf{Voorwaarde A} geldt, of als \textbf{Voorwaarde B1 en Voorwaarde B2} allebei gelden.

\prompttag{Voorwaarde A: verboden termen.}
De zin bevat een universeel ongepaste term, tenzij het gaat om een direct citaat of om expliciet uitleggen/kritiseren van schadelijk taalgebruik.

\prompttag{Voorwaarde B1: contextuele bias.}
De zin bevat stereotypering, uitsluiting, machtsongelijkheid of vooroordeel, expliciet of impliciet.

\prompttag{Voorwaarde B2: gericht op een groep.}
De bias is gericht op een specifieke sociale groep die expliciet genoemd of ondubbelzinnig geïmpliceerd is.

\tcblower
\prompttag{Gebruikerstaak.}
Gebruik alleen \pfield{sentence}, \pfield{evidence} en ruwe \pfield{jury\_outputs}. Geef voorrang aan geciteerde taalsignalen en consistente multi-agent signalen boven vage indrukken; dit is geen gewone meerderheidsstemming. Gebruik confidence en risk scores alleen als weegsignalen.

\prompttag{Outputcontract.}
Retourneer exact een JSON-object:
\begin{quote}\ttfamily\footnotesize
\{ "label": 0/1, "confidence": 0--1, "rationale": "max 3 zinnen; citeer doorslaggevende signalen/agents" \}
\end{quote}
Wanneer jurors conflicteren, geef voorrang aan geciteerde verboden termen, ontmenselijkende of dreiging/last-framing, en analyses met hogere confidence alleen wanneer zij gegrond zijn in concrete signalen.
\end{dutchpromptbox}

\newpage
\begin{dutchpromptbox}{D6. Restoration Agent}
\prompttag{Systeemrol.}
Herschrijf Nederlandse beleid-achtige zinnen om taalkundige bias te verwijderen met behoud van oorspronkelijke betekenis, feiten, administratief register en beleidsintentie. Ga ervan uit dat de zin eerder in de pipeline al als biased is gemarkeerd; de taak is mitigatie, niet herclassificatie.

\prompttag{Editbeleid.}
Maak de kleinst mogelijke edits die de bias neutraliseren: vervang verboden of problematische termen, verwijder stigmatiserende metaforen, verwijder stereotypering of ontmenselijkende framing, en gebruik waar passend neutrale of person-first formuleringen. Houd de output bij een formele Nederlandse zin. Voeg geen nieuwe feiten, actoren, getallen, causale claims of aannames toe.

\tcblower
\prompttag{Gebruikerstaak.}
Gegeven \pfield{sentence}, \pfield{evidence}, \pfield{aggregate}, \pfield{orig\_label} en \pfield{sample\_id}, retourneer exact een JSON-object:
\begin{quote}\ttfamily\footnotesize
\{ "rewritten\_text": "...", "new\_label": 0, "original\_label": \pfield{orig\_label}, "id": "\pfield{sample\_id}", "notes": "1--2 kernedits" \}
\end{quote}
\promptwarn{Fidelityregel.}
Behoud alle concrete feiten uit de oorspronkelijke zin, tenzij de feitelijke formulering zelf de biased framing is. De herschrijving mag geen uitleg, bulletlijst of meerzinnige parafrase worden.
\end{dutchpromptbox}

\end{document}